\documentclass[twoside,11pt]{article}

\usepackage{jmlr2e}

\usepackage{graphicx}
\usepackage{xcolor}
\usepackage{amstext}
\usepackage{amsmath}
\usepackage{amssymb}
\usepackage{booktabs}
\usepackage{subfigure}
\newtheorem{myDef}{Definition}

\usepackage[flushleft]{threeparttable}
\usepackage{array, multirow, multicol}
\DeclareMathOperator{\tr}{tr}

\usepackage{algorithm}
\usepackage{algpseudocode}

\algnewcommand\algorithmicinput{\textbf{Input:}}
\algnewcommand\Input{\item[\algorithmicinput]}
\algnewcommand\algorithmicoutput{\textbf{Output:}}
\algnewcommand\Output{\item[\algorithmicoutput]}

\jmlrheading{X}{XXXX}{1-XX}{XX/XX}{XX/XX}{XXX}{Kuo Gai and Shihua Zhang}

\ShortHeadings{Tessellated Wasserstein Auto-Encoders}{Kuo Gai and Shihua Zhang}
\firstpageno{1}

\begin{document}
	
	\title{Tessellated Wasserstein Auto-Encoders}
	
	\author{\name Kuo Gai$^{1,2}$ and 
		\name Shihua Zhang$^{1,2,*}$ \\ 
		\addr $^{1}$Academy of Mathematics and Systems Science\\
		Chinese Academy of Sciences\\
		Beijing 100190, China\\
		$^{2}$School of Mathematical Sciences\\
		University of Chinese Academy of Sciences\\
		Beijing 100049, China
		\\ $^*$Email: zsh@amss.ac.cn
	}

\maketitle

\begin{abstract}
Non-adversarial generative models are relatively easy to train and have less mode collapse compared to adversarial ones. However, they are not very accurate in approximating the target distribution in the latent space because they don't have a discriminator. To this end, we develop a novel non-adversarial framework called Tessellated Wasserstein Auto-Encoders (TWAE) to tessellate the support of the target distribution into a given number of regions by the centroidal Voronoi tessellation (CVT) technique and design batches of data according to the tessellation instead of random shuffling for accurate computation of discrepancy. Theoretically, we demonstrate that the error of estimate to the discrepancy decreases when the numbers of samples $n$ and regions $m$ of the tessellation become larger with rates of $\mathcal{O}(\frac{1}{\sqrt{n}})$ and $\mathcal{O}(\frac{1}{\sqrt{m}})$, respectively. Given fixed $n$ and $m$, a necessary condition for the upper bound of measurement error to be minimized is that the tessellation is the one determined by CVT. TWAE is very flexible to different non-adversarial metrics and can substantially enhance their generative performance in terms of Fr\'{e}chet inception distance (FID) compared to existing ones. Moreover, numerical results indeed demonstrate that TWAE is competitive to the adversarial model, demonstrating its powerful generative ability.
\end{abstract}


\section{Introduction}\label{sec:introduction}
Knowing the distribution of data is a fundamental task of data science. Prior distributions such as Laplacian, Gaussian and Gaussian mixture distributions are often used to model the data. However, their ability of representation is limited. With the rise of deep learning, we can use more parameters to model the distribution accurately. The basic assumption of such methods is that complex high-dimensional data such as images concentrate near a low-dimensional manifold. Generative adversarial network (GAN) \citep{goodfellow2014generative} and Wasserstein auto-encoder with generative adversarial network (WAE-GAN) (also known as adversarial auto-encoder (AAE)) \citep{makhzani2015adversarial,tolstikhin2017wasserstein} are the representatives and have many variants. GAN trains a generator to generate new samples and a discriminator to teach the generator to improve its quality. From a probabilistic view, the generator maps points from a simple low-dimensional distribution such as a uniform distribution or a Gaussian distribution to the target high-dimensional distribution (e.g., face or handwriting images), while the discriminator computes the discrepancy between the generated distribution and the target one. WAE-GAN trains an invertible mapping between two distributions with the Wasserstein distance as the reconstruction loss, i.e., an encoder from the data space to the latent space and a decoder from the latent space to the data space. WAE-GAN employs GAN to minimize the discrepancy between the output of the encoder and the samplable prior distribution in the latent space. Both methods use adversarial training, i.e., a two player game between generator (encoder) and discriminator.

\par As we know that GAN is hard to train.  \citet{arjovsky2017wasserstein,arjovsky2017towards} ascribed this to the choice of discrepancy. Classifical GAN uses KL-divergence and performs good under some tricks \citep{salimans2016improved}. But in theory, when the supports of two distributions are disjoint, KL-divergence fails and causes unstability of the model. A more stable variant Wasserstein-GAN (WGAN) introduced from the optimal transport view uses a discriminator with clipped parameters to compute the Wasserstein distance. However, clipping limits the discriminator to find the subtle difference between two distributions. Another strategy imposes the one-Lipschitz constraint by regularization methods. Since the Wasserstein distance is a real distance, the optimization appears more stable and converges faster than GAN. Apart from the optimal transport, several other studies have also been proposed to explain and improve this \citep[][]{salimans2016improved,miyato2018spectral,isola2017image}.

\par  The complexity of high-dimensional data and the instability of adversarial models lead to mode collapse, which is the main obstacle for GANs in many applications. The mode collapse in GANs refers to the problem of overfitting to a part of the training modes and forget the rest. \citet{lucic2018gans} showed that even the best GAN dropped $72\%$ of the modes. In theory, \citet{arora2017generalization} proved that the trained distribution can not converge to the target one with several standard metrics. This can be blamed on the adversarial mechanism. In game theory, based on gradient descent optimization algorithm, the discriminator and generator find a local Nash equilibrium rather than a global one. From a statistical view, the discriminator has cumulative preference of mode when it classifies real and fake data in the training process, since the discriminator is trained based on the former step. So the discriminator is sensitive to some modes and insensitive to others. More formally, the estimation of discrepancy is biased, which makes the generated distribution not converge to the target one.

\par To solve this problem, a potential approach is to find alternatives of the adversarial mechanism by computing the discrepancy without neural network for discrimination. For example, a kernel-based method maximum mean discrepancy (MMD) shows a good property on approximating the independent and identically distributed (i.i.d.) Gaussian distribution and finds its usage on WAE-MMD \citep{tolstikhin2017wasserstein} and MMD-GAN \citep{li2017mmd}. However, MMD only matches principle features of two distributions and lose other ones which cannot be captured by the kernel. As to the discrepancy of arbitrary distributions, researchers have introduced a new metric called the sliced-Wasserstein (SW) distance \citep{bonnotte2013unidimensional}, which has similar qualitative properties with the Wasserstein distance. But it is much easier to compute. Inspired by the one-dimensional case of the Wasserstein distance, the data is projected onto an one-dimensional subspace for analytical solution, then the SW distance is obtained by integrating over all the subspaces.
Thus, the number of samples needed to estimate the integration increases as the dimension of data goes up. More generally, the SW distance has been generalized to the high-dimensional situation, where the data is projected into a $k$-dimensional subspace ($k\geq2$), which maximizes their transport cost \citep{paty2019subspace,lin2020projection}. This distance is more robust to noise compared with the Wasserstein distance because of its dimension reduction operation.

\par Compared to adversarial training, non-adversarial approaches have no cumulative preference since they do not memorize historical information and are easy to train due to the unemployment of the discriminator. However, since the distribution of high-dimensional data concentrates near a low-dimensional manifold, where the Euclidean distance is no longer effective, non-adversarial approaches are not over-parameterized to learn the distance on the manifold. So they may be cursed by high dimensionality. This means, when the dimension is high and the shape of the manifold is complicated, the error of the estimation to the discrepancy may be beyond tolerance. As a consequence, the performance of non-adversarial algorithms such as variational auto-encoder (VAE) \citep{kingma2013auto}, WAE-MMD, sliced-Wasserstein auto-encoder (SWAE) \citep{kolouri2018sliced} are not as good as that of WAE-GAN or variants of GAN under similar architectures of neural network.
\begin{figure}[t!]
	\centering
	\includegraphics[width=0.84\columnwidth]{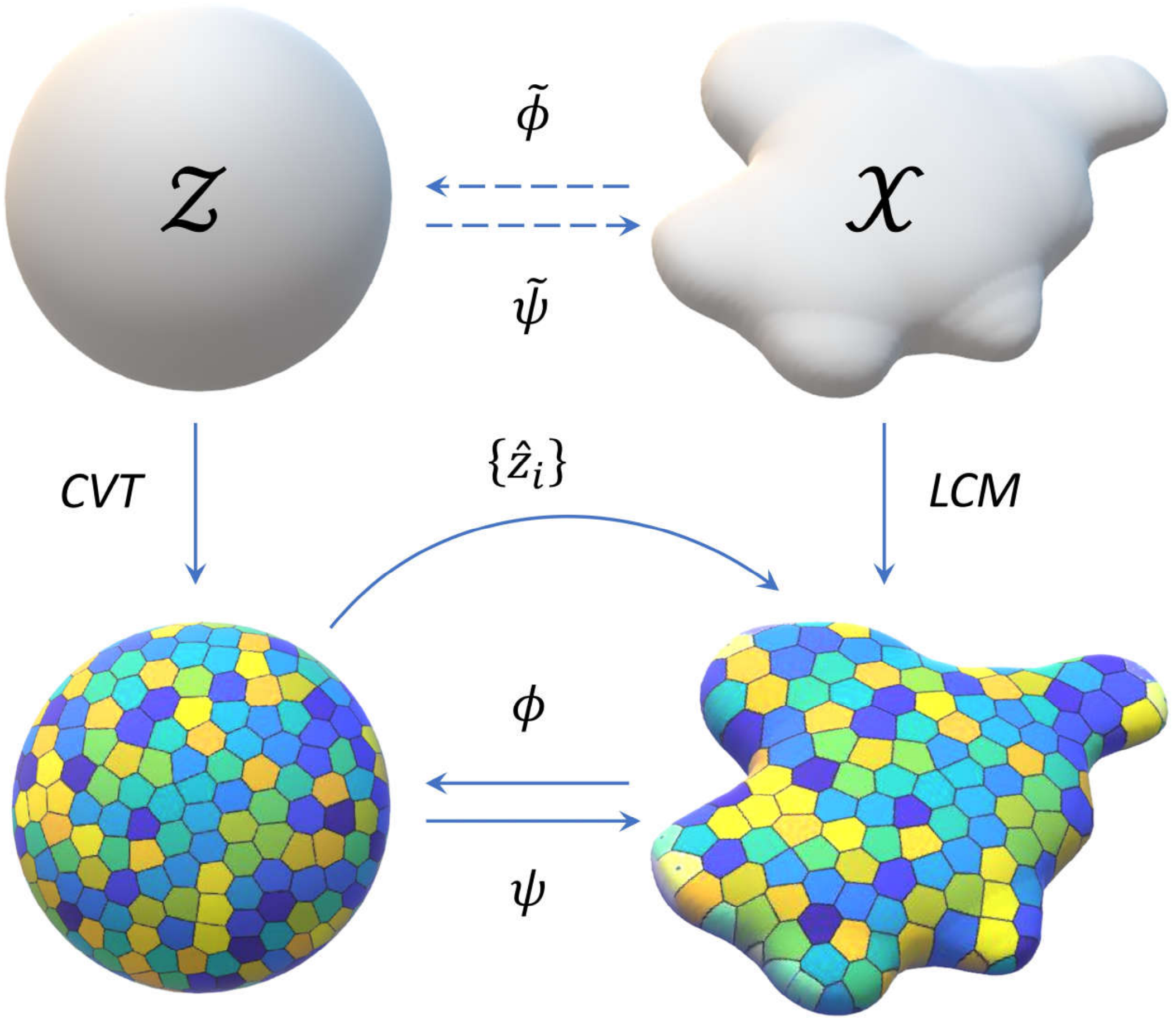}
	\caption{Illustration of TWAE. In a traditional way, the encoder $\tilde{\phi}:\mathcal{X}\to\mathcal{Z}$ and decoder $\tilde{\psi}:\mathcal{Z}\to\mathcal{X}$ are trained using randomly shuffled batches of data. In TWAE, the support of a known distribution is tessellated by the centroidal Voronoi tessellation (CVT) procedure, then the batch of data is designed with the least cost method (LCM) by their distance to the centroid $\{\widehat{z}_i\}$ of each region. In the end, the auto-encoder $(\phi,\psi)$ is trained over region by region.}
	\label{fig:1}
\end{figure}

\par In this paper, we develop a novel non-adversarial framework--Tessellated Wasserstein Auto-Encoders (TWAE) to tessellate the support of the target distribution  in the latent space into a given number of regions and design batches of data according to the tessellation instead of random shuffling. In more detail, the cost function of classical generative auto-encoders consists of the reconstruction error in the data space and the discrepancy error in the latent space. To compute the latter, TWAE separates the computation of the global discrepancy into some local ones. To do this, we need to obtain a tessellation of the support of both the target and generated distributions (Fig. \ref{fig:1}). We implement this task in two steps: first we tessellate the support of the prior distribution; second we cluster the encoded data corresponding to the tessellation. For the first step, we provide two ways to achieve the tessellation: centroidal Voronoi tessellation (CVT) and sphere packing. CVT can generate points which are the centroids of the corresponding Voronoi regions. Asymptotically speaking, all regions of the optimal CVT are congruent to a basic region. CVT can be applied to a connected set in $R^n$ with arbitrary shapes. The sphere packing approach can tessellate the space into exactly congruent regions with $E_8$-lattice in $R^8$ and Leech lattice in $R^{24}$. For the second step, we adopt an assignment algorithm to keep the correspondence of the encoded data and the regions of tessellation. Thereby the discrepancy on the whole support is separated into a sum of local discrepancies on each region. Compared with traditional ways of sampling on the whole support, TWAE can sample specially in each region. As a result, we can force the generated distribution to approximate the target one better. Since the tessellation is independent of discrepancy metrics, TWAE is compatible to different ones and enhance their performance.

\par The rest of this paper is organized as follows. In section 2, we start from the optimal transport and briefly review the optimal transport-based generative methods. To the end, we introduce CVT and sphere packing as basic tools to achieve the tessellation. In section 3, we describe TWAE in details. In section 4, we derive the sample and measurement error of TWAE theoretically. In section 5, we conduct extensive experiments to demonstrate the effectiveness of TWAE. In section 6, we provide discussion and conclusion.

\section{Related Work}
In this section, we start from the optimal transport in Sec 2.1 and briefly review the optimal transport-based generative methods such as WGAN \citep{arjovsky2017wasserstein}, sliced-Wasserstein GAN (SWGAN) \citep{deshpande2018generative}, WAE  and SWAE \citep{kolouri2018sliced} in Sec 2.2. We further introduce CVT and sphere packing as basic tools to achieve the tessellation in Sec 2.3 and 2.4 respectively.
\subsection{Optimal transport}
\par The optimal transport problem stems from a problem on transporting commodities. Suppose there are $m$ sources $x_1,\cdots,x_m$ for a commodity, with $a_i$ units of supply at $x_i$ and $n$ sinks $y_1,\cdots,y_n$ for it, with $b_i$ units of demand at $y_i$, $c_{ij}$ $(i=1,\cdots,m; j=1,\cdots,n)$ is the cost of transporting one unit of this commodity from $x_i$ to $y_j$. We wish to find a transport plan $\{f_{ij}|i=1,\cdots,m; j=1,\cdots,n\}$ to minimize the total cost. The problem can be formulated as
\begin{equation}
\begin{aligned}
\min\quad &\sum\limits_{i,j} c_{i j} f_{i j}\\
\mbox{s.t.}\quad&\sum\limits_{j=1}^n f_{i j}= a_i, \quad i=1,\cdots,m\\
&\sum\limits_{i=1}^{m} f_{i j}= b_j, \quad i=1,\cdots,n  \\
&\qquad f_{i j}\geq 0,
\end{aligned}
\label{eq:monge}
\end{equation}
which can be solved by linear programming. Since the computational complexity of solving (\ref{eq:monge}) is $\max\{m^3,n^3\}$, it can be expensive when $\max\{m,n\}$ is large. One solution for this computational problem is by using the entropic regularized version of optimal transport \citep{cuturi2013sinkhorn}. It can trade off a little optimality in exchange for an improved complexity of $\mathcal{O}\left(\max\{m^2,n^2\}/\epsilon^2\right)$, where $\epsilon>0$ stands for the accuracy level \citep{dvurechensky2018computational, lin2019efficient}, which is more scalable than solving (\ref{eq:monge}) by using linear programming. However, it is still expensive when we need to compute the optimal transport problem repeatedly, especially in learning the data distribution.

\par With the development of measure theory, the optimal transport problem can be stated as follows
\begin{equation}
W_c(P_x,P_y)=\inf_{\mathcal{T}}\mathbb{E}_{x\sim P_x}[c(x,\mathcal{T}(x))],
\label{eq:wasserstein}
\end{equation}
where $\mathcal{T}:X \to Y $ is a measure preserving transformation. This is known as the Monge formulation of the optimal transport problem \citep{villani2003topics}. There can be no admissible $\mathcal{T}$, for instance if $P_x$ is a Dirac delta and $P_y$ is not. To overcome this difficulty, \citet{kantorovich2006problem} proposed the following way to relax this problem
\begin{equation}
W_{c}\left(P_{x}, P_{y}\right) =\inf _{\Gamma \in {\Pi}\left(P_{x}, P_{y}\right)} \mathbb{E}_{(x, y) \sim \Gamma}[c(x, y)],
\end{equation}
where $\Pi(P_{x}, P_{y})$ denotes the set of all joint distributions $\Gamma(x,y)$ whose marginals are respectively $P_x$ and $P_y$. $c:X\times Y\to [0,\infty]$ is the cost function of transport. Particularly, $W_1$ and $W_2$ denote the Wasserstein distance when $c(x,y)=\|x-y\|_1$ and $c(x,y)=\|x-y\|_2^2$, respectively. Intuitively, $\Gamma(x,y)$ indicates how much ``mass" must be transported from $x$ to $y$ in order to transform the distribution $P_x$ into the distribution $P_y$. The infimum of the transport cost is called the Wasserstein distance of two distributions $P_x$ and $P_y$. The Wasserstein distance is a true distance and has a finer topology to guarantee convergence when minimize the distance. But the Wasserstein distance is hard to compute because the feasible region of $\Pi(P_{x}, P_{y})$ is too large to search. If the two distributions are assumed to be Gaussian, i.e., $x\sim \mathcal{N}(m_1,\Sigma_1)$, $y\sim\mathcal{N}(m_2,\Sigma_2)$ with the means $m_1$, $m_2 \in \mathbb{R}^p$ and the covariance $\Sigma_1$, $\Sigma_2 \in \mathbb{R}^{p\times p}$, their squared Wasserstein distance has a closed form \citep[][]{olkin1982distance}
\begin{equation}
GW^2=W_{2}^2\left(P_x, P_y\right)=\left\|m_{1}-m_{2}\right\|_{2}^{2}+\operatorname{tr}\left(\Sigma_{1}+\Sigma_{2}-2\left(\Sigma_{2}^{1 / 2} \Sigma_{1} \Sigma_{2}^{1 / 2}\right)^{1 / 2}\right).
\label{eq:Wofgaussian}
\end{equation}
This is denoted as the GW distance.
\subsection{Limit laws of the empirical Wasserstein distance}
In practice, the size of data set $\{x_i\}_{i=1}^N$ is too large for the linear programing in (\ref{eq:monge}) and we sample batches for better computation. Let $P_N$ denote the empirical distribution of $\{x_i\}_{i=1}^N$, and $P_n$ denote the empirical distribution of $n$ i.i.d. samples from $P_N$. Thus, the speed of convergence of the Wasserstein distance $P_n$ to $P_N$ is of importance. \citet*{sommerfeld2018inference} showed that the convergence rate is $n^{-\frac{1}{2}}$, i.e.,
\begin{theorem} \label{thm:sommer}
	 With $n$ approaching infinity $$\sqrt{n}W_2^2(P_n,P_N)\to \gamma_1, $$
	where $\gamma_1$ is a random variable correlated with $P_N$.
\end{theorem}
\par This theorem indicates that the convergence rate of empirical distribution is independent of the dimension. So we need not worry about the curse of dimensionality. However, if $P$ is absolutely continuous on $R^d$, then $\mathbb{E}\left[W_2(P_n,P)\right]>Cn^{-\frac{1}{d}}$ \citep{weed2019sharp}. Since computation of the GW distance is based on mean and covariance of empirical distributions, this asymptotic property can be generalized to the GW distance.
\begin{theorem}
	Let $P\neq Q$ be Gaussian, $P\sim N(m_1,\Sigma_1)$, $Q\sim N(m_2,\Sigma_2)$ with $\Sigma_1$ and $\Sigma_2$ having full rank. Let $P_n$ and $Q_n$ be generated by i.i.d. samples $z_1,\cdots,z_n\sim Q$ and $\tilde{z}_1,\cdots,\tilde{z}_n\sim P$, respectively. $P_n^{'}$ is an independent copy of $P_n$. Then with $n$ approaching infinity
	$$
	\sqrt{n}\left(G W^{2}\left(P_{n}, Q_{n}\right)-G W^{2}(P, Q)\right) \rightarrow N(0, w),
	$$
	$$
	nGW^2(P_n,P_n^{'})\to \gamma_2,
	$$
	where $w$ is correlated with $P$ and $Q$, and $\gamma_2$ is correlated with $P$.
	\label{thm:Gaussian}
\end{theorem}
\par When the dimension of distributions is one, similar results hold for continuous distributions under moderate conditions \citep{bobkov2014one}.
\begin{theorem}\label{thm: onedim}
	Let $P$, $Q$ be continuous distributions on $\mathbb{R}$. Let $P_n$ and $Q_n$ be generated by i.i.d. samples $z_1,\cdots,z_n\sim Q$ and $\tilde{z}_1,\cdots,\tilde{z}_n\sim P$ respectively. $P_n^{'}$ is an independent copy of $P_n$. Let
	\begin{equation}
	J_2(P)=\int_{F^{-1}(0)}^{F^{-1}(1)}\frac{F(x)\left(1-F(x)\right)}{p(x)}dx,
	\end{equation}
	where $p$ is the density of distribution $P$ and $F$ is the cumulative function of $P$.
	 If $\max\{J_2(P),J_2(Q)\}<\infty$, then with $n$ approaching infinity
	$$
	\sqrt{n}\left(W_2^2\left(P_n,Q_n\right)-W_2^2\left(P,Q\right)\right)\to N(0,\sigma^2),
	$$
	$$
	nW_2^2\left(P_n,P_n^{'}\right)\to \gamma_3,
	$$
	where $\sigma^2$ is the variance correlated with $P$ and $Q$, and $\gamma_3$ is a random variable correlated with $P$.
\end{theorem}
\subsection{Optimal transport-based generative models}
\par \citet{arjovsky2017wasserstein} first approached the problem of generative modeling from the optimal transport view. The infimum in (\ref{eq:wasserstein}) is highly intractable. On the other hand, when $c(x,y)=\|x-y\|_1$, the Kantorovich-Rubinstein duality \citep{villani2008optimal} tells us that
\begin{equation}
\begin{aligned}
W_1(P_x,P_y)&=\inf_{\Gamma \in {\Pi}\left(P_{x}, P_{y}\right)}\mathbb{E}_{(x,y)\sim\Gamma}[\|x-y\|_1]\\
&=\sup_{\|f\|_L \leq1}\left(\mathbb{E}_{x\sim P_x}[f(x)]-\mathbb{E}_{y\sim P_y}[f(y)]\right),
\end{aligned}
\end{equation}
where the supremum is over all the one-Lipschitz functions $\{f:\mathcal{X}\to\mathbb{R}\}$. The function $f$ is approximated by a parameterized family of functions $\{f_w\}_{w\in \mathcal{W}}$, where $\mathcal{W}$ is the parameter space. \citet{arjovsky2017wasserstein} suggested to impose the one-Lipschitz constraint to force parameters $w$ lie in a compact space by clipping the weights to a fixed box. \citet{gulrajani2017improved} introduced a soft version of the constraint with a penalty on the gradient norm for random samples by optimizing
\begin{equation}
L=\mathbb{E}_{x\sim P_x}[f(x)]-\mathbb{E}_{y\sim P_y}[f(y)]+\lambda\mathbb{E}_{\widehat{x}\sim P_{\widehat{x}}}[(\|\nabla_{\widehat{x}}f(\widehat{x}) \|_2-1)^2].
\end{equation}

\par To improve the stability of WGAN, \citet{deshpande2018generative} developed a mechanism based on random projections as an alternative to the black-box discriminator. Notice that the squared Wasserstein distance of two one-dimensional distributions $P_x$ and $P_y$ can be estimated accurately by sorting their samples. Suppose $x_i$, $y_i$ $(i=1,\cdots,N)$ are independently sampled from $P_x$ and $P_y$, and $x_i\leq x_{i+1}$, $y_i\leq y_{i+1}$ for all $i\in\{1,\cdots,N-1\}$, then
\begin{equation}
W_2^2(P_x,P_y)=\inf_{\Gamma \in {\Pi}\left(P_{x}, P_{y}\right)}\mathbb{E}_{(x,y)\sim\Gamma}[\|x-y\|_2^2] \approx \frac{1}{N}\sum_{i=1}^{N}(x_i-y_i)^2.
\end{equation}
Generally, if $P_x$ and $P_y$ are $d$-dimensional distributions, we project the sampled $d$-dimensional points onto one-dimensional spaces spanned by directions $w$ and integrate over all possible directions $w$ on the unit sphere $S^{d-1}$. Then we obtain the SW distance
\begin{equation}
SW_2^2(P_x,P_y)=\int_{w\in S^{d-1}}W_2^2(P_{x|w},P_{y|w})dw.
\label{eq:slicedwasserstein}
\end{equation}
Hereby $P_{x|w}$ and $P_{y|w}$ denote the projected distributions on the subspace spanned by $w$. The SW distance is a real distance and is equivalent to the Wasserstein distance as the following property holds \citep{bonnotte2013unidimensional}
\begin{equation}
SW_2^2(P_x,P_y)\leq C_{d}W_2^2(P_x,P_y)\leq C_{d}R^{\frac{1}{(d+1)}}SW_2^{\frac{1}{(d+1)}}(P_x,P_y),
\label{eq:equivalence}
\end{equation}
where $C_{d}>0$ is a constant correlated with the dimension $d$, and $P_x$, $P_y\in \mathcal{P}(B(0,R))$, where $B(0,R)$ is the ball with radius $R$ and the origin as the center point, $\mathcal{P}(\cdot)$ is the space of probability measure. The SW distance can be regarded as a good alternative to the Wasserstein distance because it can be easily acquired by random projections. However, since the area of a sphere with a radius of $r$ in $\mathbb{R}^d$ is proportional to $r^{d-1}$, the number of projections goes up exponentially with the dimension of data. Hence, the huge computation caused by the curse of dimensionality becomes a main obstacle to put it into practice. The SW-based methods sacrifice accuracy to the discrepancy for the privilege of stability without the discriminator.
\par Alternatively, \citet{deshpande2019max} proposed the max sliced-Wasserstein distance (Max-SW) to distinguish the probability distribution using only one important direction. Even though its performance in GANs is better than the SW distance, it can miss some important differences between two distributions in high-dimensional space. \citet{nguyen2020distributional} proposed distributional sliced-Wasserstein distance (DSW) to search for an optimal distribution of important directions. The DSW distance has much better performance than the SW distance for GANs while has similar computational time as the SW distance. \citet{kolouri2019generalized} proposed the generalized sliced-Wasserstein distance (GSW) to replace the linear projections in the SW distance with non-linear projections, which can fit the manifold of data better. They also suggested to generalize the Max-SW distance to the Max-GSW distance by using a single projection as long as it leads to a space with the maximal distance. With a neural network as the non-linear projection function, minimizing Max-GSW between two distributions is analogical to adversarial learning, where the goal of the adversarial network is to distinguish the two distributions. One can also solve the optimal transport problem in generative models through solving the Monge-Amp$\grave{e}$re equation. This equation can be linearized to the McKean-Vlasov equation and numerically solved using the forward Euler iteration \citep{gao2020learning}. Furthermore, \citet{lei2020geometric} proposed a variational approach named AE-OT to solve the discrete Monge-Amp$\grave{e}$re equation explicitly. AE-OT separates the computation of OT from the training of neural network and improves the transparency of generative models.

\par Another main stream of generative models is based on auto-encoders. Different from GANs, generative auto-encoders approximate a prior distribution in the latent space. Their generalized formulation is as follows
\begin{equation}
\min_{\phi,\psi} \mathbb{E}_{x\sim P_x}[c(x,\psi(\phi(x)))] +\lambda D(P_z||Q_z),
\label{eq:autoencoder}
\end{equation}
where $\phi$ is the encoder, $\psi$ is the decoder, $P_x$ is the data distribution, $P_z$ is a prior samplable distribution, $Q_z$ is the empirical distribution of the encoded data $z= \phi(x)$, and $\lambda$ indicates the relative importance of the discrepancy. In WAE \citep{tolstikhin2017wasserstein}, GAN and MMD have been proposed (denoted as WAE-GAN and WAE-MMD respectively). In SWAE \citep{kolouri2018sliced}, the choice of $D$ in (\ref{eq:autoencoder}) is the SW distance.

\subsection{Centroidal Voronoi Tessellation}
\par Given an open set $\Omega\subseteq \mathbb{R}^d$, the set $\{V_i\}_{i=1}^k$ is called a tessellation of $\Omega$ if $V_i\cap V_j=\varnothing$ for $i\neq j$ and $\cup_{i=1}^k \overline{V}_i=\overline{\Omega}$ ($\overline{\Omega}$ means the closed hull of set $\Omega$). Given a set of points $\{\widehat{z}_i\}_{i=1}^k$ belonging to $\overline{\Omega}$, the set $\{\widehat{V}_i\}_{i=1}^k$ is called a Voronoi tessellation if the Voronoi region $\widehat{V}_i$ corresponding to the point $\widehat{z}_i$ is defined by
\begin{equation}
\widehat{V}_i = \{x\in \Omega | \|x-\widehat{z}_i\|<\|x-\widehat{z}_j\|\ \text{for } j=1,\cdots,k,j\neq i   \}.
\end{equation}
The points $\{\widehat{z}_i\}_{i=1}^k$ are called generators. In the rest of this paper, without special mention, a generator denotes the generator of tessellation rather than that of GAN.  Given a region $V\subseteq \mathbb{R}^d$ and a density function $\rho$, the mass centroid $z^*$ of $V$ is defined by
\begin{equation}
z^*=\frac{\int_V y\rho(y)dy}{\int_V \rho(y)dy}.
\end{equation}
If $\widehat{z}_i=z_i^*$, $i=1,\cdots,k$, i.e., the mass centroid of the region is exactly the generator, we call such a tessellation a CVT \citep{du1999centroidal}.

\par Next, we introduce the classical Lloyd's method to construct an approximate CVT in the following steps: \textbf{Step 0}: Select an initial set of $k$ points $\{z_i\}_{i=1}^k$ using a sampling strategy (e.g., Monte Carlo sampling); \textbf{Step 1}: Construct the Voronoi tessellation $\{V_i\}_{i=1}^k$ of $\Omega$ associated with the points $\{z_i\}_{i=1}^k$; \textbf{Step 2}: Compute the mass centroids of the Voronoi regions $\{V_i\}_{i=1}^k$ found in \textbf{Step 1}; these centroids are the new set of points $\{z_i\}_{i=1}^k$; \textbf{Step 3}: If this new set of points meets some convergence criteria, then terminate; otherwise, return to \textbf{Step 1}. The Lloyd's method can be viewed as an alternative iteration between the Voronoi tessellation construction and centroid computation. Clearly, a CVT is a fixed point of the iteration. If we define a clustering energy by
\begin{equation}
\mathbb{K}(\{\widehat{z}_i\}_{i=1}^k, \{\widehat{V}_i\}_{i=1}^k)=\sum_{i=1}^{k} \int_{\widehat{V}_{i}} \rho(y)\left\|y-\hat{z}_{i}\right\|^{2} d y,
\label{energyfun}
\end{equation}
then the energy associated with the Voronoi tessellation deceases monotonically during the Lloyd iterations until a CVT is reached \citep{du1999centroidal}. Apart from the Lloyd method, there is another simple one called K-means method (also known as probabilistic Lloyd method), which relies very little on the geometric information. The K-means method is defined as follows: \textbf{Step 0:} select an initial set of $k$ points $\{z_i\}_{i=1}^k$, e.g., by using a Monte Carlo method; \textbf{Step 1:} select a $y\in \Omega$ at random, according to the probability density function $\rho(y)$; \textbf{Step 2:} find the $z_i$ that is closest to $y$, and denote the index of that $z_i$ by $i^{*}$; \textbf{Step 3:} set $z_{i^{*}} \leftarrow \frac{j_{i^{*}} \cdot z_{i^{*}}+y}{j_{i^{*}}+1}$ and $j_{i^{*}}\leftarrow j_{i^{*}}+1$, then this new $z_{i^{*}}$ along with the unchanged $z_i$, $i\neq i^{*}$, forms the new set of points ${z_i}_{i=1}^k$; \textbf{Step 4:}	If this new set of points meets some convergence criteria, terminate; otherwise, go back to \textbf{Step 1}.
\par The K-means method has been analyzed in \citep{macqueen1967some}, where the almost sure convergence of energy is proved. Though attractive due to its simplicity, the convergence of the K-means method is very slow \citep{du2002numerical}. Nevertheless, the algorithm is highly amenable to fully scalable parallelization, as demonstrated in \citep{ju2002probabilistic}.

\subsection{Sphere Packing}
The CVT technique is an approximate method. In mathematics, there is an exact method based on sphere packing to tessellate the space. The standard packing problem is how to arrange spheres of equal radius to fill space as densely as possible in $R^n$. It is very hard to construct a packing scheme for an arbitrary $n$. Luckily, for the special cases, it has been proved that $E_8$-lattice ($n=8$) and Leech lattice ($n=24$) give the densest lattice packing \citep[][]{cohn2017sphere}. For $E_8$-lattice, each lattice point has 240 nearest neighbors, and for Leech lattice the number is 196560 which is too large for our tessellation considering the sizes of common data. In more detail, for $E_8$-lattice, the nearest neighbors of the origin have the shape $(\pm 1^2,0^6)$ ($2^2C^2_8=112$ of these) and $(\pm \frac{1}{2}^8)$ with even number of negative signs ($2^7=128$ of these). The set of neighbors $\Delta$ is actually the root lattice of $E_8$-lattice since $E_8=\mathbb{Z}\Delta$.
\par Though $E_8$ gives the densest packing in $\mathbb{R}^8$, it may not be optimal restricted to a region with a fixed shape. Nevertheless, for a ball $B$ in $\mathbb{R}^8$, a possible tessellation scheme utilizing $E_8$-lattice is that one point locates at the center of $B$, surrounded by 240 points in the way of $E_8$ within $B$. By adjusting the radius of packed spheres, we obtain a tessellation for $B$, which is symmetrical and has regions with exactly the same volume. Then if we tessellate the space with the tangent plane of each two spheres, we separate the space into regions with exactly the same volume rather than roughly equal one in a CVT.

\section{TWAE}
\par In this section, we follow the generalized formulation of generative auto-encoder with a reconstruction error in the data space and a discrepancy error in the latent space,
\begin{equation}
\min_{\phi,\psi} \mathbb{E}_{x\sim P_x}[c(x,\psi(\phi(x)))] +\lambda D(P_z||Q_z).
\end{equation}
In Sec 3.1, to compute the discrepancy of $P_z$ and $Q_z$ more accurately, we first derive TWAE by tessellating the support of $P_z$ and $Q_z$ simultaneously.
We further develop a new optimization strategy with non-identical batches as well as a regularizer to get better solutions in Sec 3.2.

\subsection{Model Construction}
In this paper, we propose $P_z$ to be a uniform distribution in a unit ball, then the probability of a region is proportional to its volume. We adopt the Wasserstein distance as the divergence $D$ for its good property though our tessellation framework is also flexible to other discrepancy metrics.

\par Let's go back to the discrete Wasserstein distance (\ref{eq:monge}). Suppose there are $N$ points of $\tilde{z}_i$ sampled from the prior distribution $P_z$ and the same number of $z_i$ encoded by the encoder $\phi$. $P_N$ and $Q_N$ are the empirical distribution of $\{\tilde{z}_i\}_{i=1}^N$ and $\{z_i\}_{i=1}^N$, respectively. We can compute the Wasserstein distance by assigning each $z_i$ to a $\tilde{z}_{\sigma_i}$ as follows
\begin{equation}
W(P_N,Q_N)=\frac{1}{N}\min_{\sigma} \sum_{i=1}^N \|z_i-\tilde{z}_{\sigma_i}\|,
\label{eq:discrete_W}
\end{equation}
where $\sigma$ is a permutation of an index set $\{1,\cdots,N\}$. It can be formulated as an assignment problem and solved by mature linear programming algorithms with a computational complexity of $O(N^{2.5}\mbox{log}(N))$. Sinkhorn divergence can be a good alternative with a computational complexity of $\mathcal{O}(N^2/\epsilon^2)$, where $\epsilon$ stands for the accuracy of approximation \citep{genevay2018learning}. However, when  $\epsilon$ is small, this complexity is still prohibitive for the usage in the inner loop of a learning algorithm. As mentioned before, instead of linear programming, inaccurate approaches such as clipped networks \citep[][]{arjovsky2017wasserstein} and random projection \citep[][]{deshpande2018generative} have been proposed to address it. For large $N$, the traditional way is to divide the dataset into batches and to optimize the objective function batch by batch in a gradient descent manner, which is the well-known stochastic gradient descent. However, batches with small size lose some information to model the distribution delicately. To address this issue, we combine the assignment method and the batch optimization to a two-step algorithm. That is we first design the batches according to their similarity and then minimize the discrepancy based on the optimization per batch.

\par For the first step, we find $m$ points $\{\widehat{z}_j\}_{j=1}^m$ on the support of $P_z$. $\{\widehat{z}_j\}_{j=1}^m$ can be treated as generators of a tessellation $\{V_j\}_{j=1}^m$ on the support $\Omega$ that $V_i\cap V_j=\varnothing$ for $i\neq j$ and $\cup_{i=1}^k \overline{V}_i=\overline{\Omega}$ . We assume that the volume of each $V_j$ is equal so that we can sample a batch with the same number $n$ of points in each $V_j$ to model the distribution of $P_z$ restricted on $V_j$. Assigning each encoded data point $z_i$ to one of the generators $\{\widehat{z}_j\}_{j=1}^m$ is an easier task than (\ref{eq:discrete_W}) because $m$ is much smaller than $N$. Each of $\{\widehat{z}_j\}_{j=1}^m$ is assigned by $n=\frac{N}{m}$ points. The problem can be formulated as
\begin{equation}
\begin{aligned}
\min\quad &\sum\limits_{i,j} \|z_i-\widehat{z}_j\|_2^2 f_{i j}\\
\mbox{s.t.}\quad &\sum\limits_{j=1}^m f_{i j}= 1, \  i=1,\cdots,N\\
&\sum\limits_{i=1}^{N} f_{i j}= n, \ j=1,\cdots,m  \\
&\quad f_{i j}\in \{0,1\}.\\
\end{aligned}
\label{eq:hitchcock}
\end{equation}
It is a special case of the Hitchcock problem as both the demands and supplies are equal. By doing this, the dataset $\{z_i\}_{i=1}^N$ is clustered into $m$ sets $\{S_j\}_{j=1}^m$ according to their distance to the generators $\{\widehat{z}_j\}_{j=1}^m$. Then for each cluster $S_j$ corresponding to the generator $\widehat{z}_j$, we can estimate the Wasserstein distance of $Q_z$ and $P_z$ restricted on the region $V_j$.

\par The overall discrepancy is obtained by computing the local ones upon all the sets $\{S_j\}_{j=1}^m$. Thus, we have
\begin{align}
\mathbb{E}\left[W_2^2(P_N,Q_N)\right]
&=\frac{1}{N}\mathbb{E}\left[\min_{\sigma}\sum_{i=1}^N\|z_i-\tilde{z}_{\sigma_i}\|_2^2\right]\label{eq:17}\\
&=\frac{1}{N}\mathbb{E}\left[\min_{\sigma}\sum_{j=1}^m\sum_{z_i\in S_j}\|z_i-\tilde{z}_{\sigma_i}\|_2^2\right] \label{eq:oplan}\\
&\leq \frac{1}{N}\mathbb{E}\left[\min_{\sigma}\sum_{j=1}^m\sum_{\begin{scriptsize}\begin{array}{l}
	z_i\in S_j\\
	\tilde{z}_{\sigma_i^{j}}\in V_j
	\end{array}\end{scriptsize} }\|z_i-\tilde{z}_{\sigma_i}\|_2^2\right]\label{eq:19}\\
\end{align}
\begin{align}
&=\frac{1}{N}\mathbb{E}\left[\sum_{j=1}^m\min_{\sigma^{j}}\sum_{\begin{scriptsize}\begin{array}{l}
	z_i\in S_j\\
	\tilde{z}_{\sigma_i^{j}}\in V_j
	\end{array}\end{scriptsize} }\|z_i-\tilde{z}_{\sigma_i^{j}}\|_2^2\right] \label{eq:20}\\
&=\frac{1}{m}\mathbb{E}\left[\sum_{j=1}^{m}W_2^2(P_{n|V_j},Q_{n|S_j})\right],\label{eq:tessellate}
\end{align}
where $P_{n|V_j}$ denotes the empirical distribution of $n$ samples of $P_z$ restricted on $V_j$, $Q_{n|S_j}$ denotes the empirical distribution of $S_j$, $\sigma^{j}$ denotes a permutation of an index set $\{1,\cdots,n\}$ corresponding to the region $V_j$. The inequality in (\ref{eq:19}) is because the right side has more restriction that $\tilde{z}_{\sigma_i^{j}}\in V_j$. The equality in (\ref{eq:20}) is because $z_i$ and $\tilde{z}_{\sigma_i^{j}}$ are restricted to $S_j$ and $V_j$ respectively.
   When $P_z=Q_z$, since $S_j$ is a set of points which are the closest to $\widehat{z}_j$ and $\{V_j\}_{j=1}^m$ is CVT, then for a fixed $z_i\in S_j$, its optimal match $\tilde{z}_{\sigma_i}$ in (\ref{eq:oplan}) belongs to $V_j$ with high probability. If we fix $m$ and let $N$ approach infinity, the equality holds in (\ref{eq:19}). We assume that in the training procedure, $N\gg m$ and after a few iterations, $Q_z$ and $P_z$ are approximately equal so that we can optimize the subproblems on the right side of (\ref{eq:tessellate}) instead.

\par We expect the sum of errors of estimates to the local discrepancies is smaller than the error on the whole support with the same estimator. We assume the total error can be divided into measurement error $e_m$ and sampling error $e_s$. First, the measurement error denotes the error of the estimated Wasserstein distance. In general, the measurement error is a high-level minim of the true discrete Wasserstein distance. As the sum of estimations on the regions is almost equal to that on the whole support, the sum of measurement errors ($e_m$) on regions should be smaller. Second, traditionally, we sample a batch of points from the whole distribution, so fewer points locate in a region of the support. Now we sample a batch in a local region to find the more subtle discrepancy and approximate the prior distribution better. Thus, the sampling error in local regions ($e_s$) is smaller. Our main results are that $e_m$ and $e_s$ decrease with rates of $\mathcal{O}(\frac{1}{\sqrt{m}})$ and $\mathcal{O}(\frac{1}{\sqrt{n}})$, respectively. We leave it to Section 4 for detailed theoretical exploration.

\par The whole scheme of the algorithm is summarized in \textbf{Algorithm 1}. Here we adopt the CVT technique to generate a proper tessellation. We compute CVT in the unit ball of the latent space to tessellate it into $m$ regions with approximately equal volume. We follow the procedure of the Lloyd's method and minimize the energy function in (\ref{energyfun}) to obtain the generators and a CVT. The CVT we computed is empirically good though it is not guaranteed to be the global minimum. The generators are fixed in the training process of the auto-encoder. The Hitchcock problem needs to be solved in each iteration, and it still costs too much to find the optimal solution. We adopt the least cost method (LCM) instead, which is a heuristic algorithm. We find the smallest admissible item $d_{ij}^{*}$ of the distance matrix between $\{z_i\}_{i=1}^N$ and $\{\widehat{z}_j\}_{j=1}^{m}$, and assign $z_i$ to $\widehat{z}_i$ if $\widehat{z}_i$ is not saturated. The scheme of LCM is summarized in \textbf{Algorithm 2}. As to the discrepancy, we propose two non-adversarial methods based on the GW distance (\ref{eq:Wofgaussian}) and the SW distance (\ref{eq:slicedwasserstein}). Both discrepancy metrics can be computed efficiently.
\begin{table}[t!]
	\begin{minipage}{\columnwidth}
		\begin{algorithm}[H]
			\begin{algorithmic}[1]\caption{\textbf{TWAE}}\label{algo:twae}
				\Input data $\{x_i\}_{i=1}^N$, CVT generators $\{\widehat{z}_i\}_{i=1}^m$, hyperparameter $\lambda$
				\Output encoder $\phi$, decoder $\psi$
				\Repeat
				\State $z_i=\phi(x_i)$, $i\in \{1,\cdots N\}$
				\State assign $\{z_i\}$ to $\{\widehat{z}_i\}$ by Algorithm \ref{algo:least-cost} and obtain $\{S_i\}_{i=1}^m$
				\For{$k=1\to m$}
				\State sample $n$ points $\{\tilde{z}_s^k\}$ in the region $V_k$
				\State compute $\mathcal{L}_{latent}^{k}=W(P_{n|V_k},Q_{n|S_k})$
				\State $\mathcal{L}_{recons}^{k}=\sum_{x\in\{x_t|z_t\in S_k\}}\|x-\psi(\phi(x))\| $
				\State update $\phi$ and $\psi$ by minimizing $\mathcal{L}^{k}=\mathcal{L}_{recons}^{k}+\lambda\mathcal{L}_{latent}^{k}$
				\EndFor
				\Until{convergence}
				
			\end{algorithmic}
		\end{algorithm}
	\end{minipage}
\end{table}

\begin{table}[t!]
	\begin{minipage}{\columnwidth}
		\begin{algorithm}[H]
			\begin{algorithmic}[1]\caption{\textbf{LCM}}\label{algo:least-cost}
				\Input encoded data $\{z_i\}_{i=1}^N$, generators $\{\widehat{z}_i\}_{i=1}^m$,
				\Output  clusters $S_i$, $i=1,\cdots,m$
				\State compute the distant matrix $M_{N\times m}$
				\State $S_i=\varnothing$, $i=1,\cdots,m$
				\Repeat
				\State find the minimum item $d_{ij}$ in $M$
				\State $S_j=S_j\cup\{z_i\}$
				\State mask the $i_{th}$ row in $M$
				\If{$|S_j|=n$}
				\State mask the $j_{th}$ column in $M$
				\EndIf
				\Until{all of $\{z_i\}$ is assigned}
				
			\end{algorithmic}
		\end{algorithm}
	\end{minipage}
\end{table}

\subsection{Optimization with Non-identical Batches}
In TWAE, the data points are separated into different batches according to their corresponding encoded representations in the latent space. Here $f(\theta)$ denotes the loss function of TWAE, $\theta$ denotes the parameters in the encoder and decoder, i.e.,
\begin{equation}
f(\theta)=\mathbb{E}_{x \sim P_{x}}[c(x, \psi(\phi(x)))]+\lambda D\left(P_{z} \| Q_{z}\right).
\end{equation}
Let $f_i(\theta)$ denote the loss function corresponding to the $i$-th batch of data. Thus, we have
\begin{equation}
f(\theta)=\sum_{i=1}^{m} f_{i}(\theta),
\end{equation}
where $m$ is the number of batches. In the setting of TWAE, since $f_i(\theta)$ and $f_j(\theta)\ (i\neq j)$ correspond to batches with different distributions, the value of $f_i(\theta)$ may increase with the decrease of $f_j(\theta)$ . This can result in instability for the autoencoder when it is optimized batch by batch. To solve this problem, we adopt a new optimization method attempting to keep the value of $f_i(\theta)\ (i\neq j)$ non-increasing when we optimize with the $j$-th batch of data.

\par Consider the first-order Taylor expansion of $f_i(\theta)$ with respect to $\bar{\theta}$
\begin{equation}
f_{i}(\theta)=f_{i}^{(1)}(\theta)+R_{i}(\theta),
\end{equation}
where $f_{i}^{(1)}(\theta)=f_{i}(\bar{\theta})+\nabla f_{i}(\bar{\theta})(\theta-\bar{\theta})$ and $R_{i}(\theta)=f_{i}(\theta)-f_{i}^{(1)}(\theta)$. Note that the optimization can be effective with the i.i.d. batches, which have similar loss function values. Inspired by this, to strengthen the similarity of $f_i(\theta)$  and  $f_j(\theta)\ (i\neq j)$   in TWAE, we replace $R_i(\theta)$  and $R_j(\theta)$  to $R(\theta)$ , where $R(\theta)=f(\theta)-f^{(1)}(\theta)$ . Finally, the loss function of the $i$-th batch is $f^{(1)}_i(\theta)+\alpha R(\theta)$ , where $\alpha$  is a hyper-parameter to balance the two terms.
For the $k$-th iteration (corresponding to the $k$-th batch), the value of parameters is denoted by $\theta_k$ . The Taylor series of $f_k(\theta_k)$ is expanded with respect to $\theta_{k-1}$. Thus the gradient of $f_k^{(1)}(\theta_k)+\alpha R(\theta_k)$ is
\begin{equation}
\begin{aligned}
& \nabla_{\theta_{k}}\left(f_{k}^{(1)}\left(\theta_{k}\right)+\alpha R\left(\theta_{k}\right)\right) \\
=& \nabla_{\theta_{k}}\left(f_{k}\left(\theta_{k-1}\right)+\nabla_{\theta_{k-1}} f_{k}\left(\theta_{k-1}\right)\left(\theta_{k}-\theta_{k-1}\right)\right)+\alpha \nabla_{\theta_{k}}\left(f\left(\theta_{k}\right)-f^{(1)}\left(\theta_{k}\right)\right) \\
=& \nabla_{\theta_{k-1}} f_{k}\left(\theta_{k-1}\right)+\alpha \nabla_{\theta_{k}}\left(f\left(\theta_{k}\right)\right)-\alpha \nabla_{\theta_{k}}\left(f^{(1)}\left(\theta_{k}\right)\right) \\
=& \nabla_{\theta_{k-1}} f_{k}\left(\theta_{k-1}\right)+\alpha\left[\nabla_{\theta_{k}}\left(f\left(\theta_{k}\right)\right)-\nabla_{\theta_{k-1}}\left(f\left(\theta_{k-1}\right)\right)\right].
\end{aligned}
\end{equation}

In practice, it is not convenient to compute $\nabla_{\theta_{k-1}} f_{k}\left(\theta_{k-1}\right)$ with the $k$-th batch of data and parameters in the $(k-1)$-th iteration, thus we compute $\nabla_{\theta_{k}} f_{k}\left(\theta_{k}\right)$ instead. For the second term $\alpha\left[\nabla_{\theta_{k}}\left(f\left(\theta_{k}\right)\right)-\nabla_{\theta_{k-1}}\left(f\left(\theta_{k-1}\right)\right)\right]$, it is unrealistic to compute $\nabla f(\theta)$ as the number of data points is huge. Actually, the only thing that matters is the variation $\nabla_{\theta_{k}}\left(f\left(\theta_{k}\right)\right)-\nabla_{\theta_{k-1}}\left(f\left(\theta_{k-1}\right)\right)$. We estimate $\nabla_{\theta_{k}}\left(f\left(\theta_{k}\right)\right)$ and $\nabla_{\theta_{k-1}}\left(f\left(\theta_{k-1}\right)\right)$ with the same batch randomly sampled from the whole dataset for better accuracy. To illustrate this, we have the following theorem.
\begin{theorem}
Let $f_{S^{(k)}}(\theta_k)$ and $f_{S^{(k-1)}}(\theta_{k-1})$ be estimates to $f\left(\theta_{k}\right)$ and $f\left(\theta_{k-1}\right)$ with random batches $S^{(k)}$ and $S^{(k-1)}$, respectively. Assume that $f$, $f_{S^{(k)}}$ and $f_{S^{(k-1)}}$ are two-time differentiable functions. Then the estimate error to the variation $\nabla_{\theta_{k}}\left(f\left(\theta_{k}\right)\right)-\nabla_{\theta_{k-1}}\left(f\left(\theta_{k-1}\right)\right)$, i.e.,
\begin{equation}
e=\left\|\left[\nabla_{\theta_{k}}\left(f_{S^{(k)}}\left(\theta_{k}\right)\right)-\nabla_{\theta_{k-1}}\left(f_{S^{(k-1)}}\left(\theta_{k-1}\right)\right)\right]-\left[\nabla_{\theta_{k}}\left(f\left(\theta_{k}\right)\right)-\nabla_{\theta_{k-1}}\left(f\left(\theta_{k-1}\right)\right)\right]\right\|_2,
\end{equation}
is minimized when $S^{(k)}=S^{(k-1)}$.
\end{theorem}
\begin{proof}Let
	\begin{equation}
	F(S,\theta)\triangleq \nabla_{\theta}\left(f_{S}\left(\theta\right)\right)-\nabla_{\theta}\left(f\left(\theta\right)\right),
	\end{equation}
	then we have
	\begin{equation}
	\begin{aligned}
	e&=\left\|\left[\nabla_{\theta_{k}}\left(f_{S^{(k)}}\left(\theta_{k}\right)\right)-\nabla_{\theta_{k-1}}\left(f_{S^{(k-1)}}\left(\theta_{k-1}\right)\right)\right]-\left[\nabla_{\theta_{k}}\left(f\left(\theta_{k}\right)\right)-\nabla_{\theta_{k-1}}\left(f\left(\theta_{k-1}\right)\right)\right]\right\|_2\\
	&=\left\|\left[\nabla_{\theta_{k}}\left(f_{S^{(k)}}\left(\theta_{k}\right)\right)-\nabla_{\theta_{k}}\left(f\left(\theta_{k}\right)\right)\right]-[    \nabla_{\theta_{k-1}}\left(f_{S^{(k-1)}}\left(\theta_{k-1}\right)\right)-\nabla_{\theta_{k-1}}\left(f\left(\theta_{k-1}\right)\right)]\right\|_2\\
	&=\left\|F\left(S^{(k)},\theta_{k}\right)-F\left(S^{(k-1)},\theta_{k-1}\right)\right\|_2\\
	&=\mathcal{O}(\|S^{(k)}-S^{(k-1)}\|_2)+\mathcal{O}(\|\theta_{k}-\theta_{k-1}\|_2).
	\end{aligned}
	\end{equation}
	The last equality is obtained by taking the first-order Taylor expansion of $F\left(S^{(k)},\theta_{k}\right)$ with respect to $S^{(k-1)}$ and $\theta_{k-1}$. Then $e$ is minimized when $S^{(k)}=S^{(k-1)}$.
\end{proof}

\par To conclude, the gradient in each iteration is computed with respect to two batches of data, i.e., one batch is restricted in a region for $\nabla_{\theta_{k}} f_{k}\left(\theta_{k}\right)$ and another batch is sampled from the whole support for $\nabla_{\theta_{k}}\left(f\left(\theta_{k}\right)\right)-\nabla_{\theta_{k-1}}\left(f\left(\theta_{k-1}\right)\right)$. This optimization strategy is inspired by CEASE \citep{fan2019communication} and CSL \citep{jordan2019communication} algorithms in distributed computing, where $f_i^{(1)}(\theta)$ is changed into $f^{(1)}(\theta)$ in each node machine under the assumption that data in different node machines are identically distributed. On the contrary, we assume the supports of distributions in different batches are disjoint, so we keep the first-order Taylor expansion unchanged to retain the differences. The algorithm of TWAE with regularization is summarized in \textbf{Algorithm \ref{algo:twae-r}}.

\begin{table}
	\begin{minipage}{\columnwidth}
		\begin{algorithm}[H]
			\begin{algorithmic}[1]\caption{\textbf{TWAE with regularization}}\label{algo:twae-r}
				\Input data $\{x_i\}_{i=1}^N$, CVT generators $\{\widehat{z}_i\}_{i=1}^m$, hyperparameter $\lambda$, $\alpha$, learning rate $\gamma$
				\Output encoder $\phi$, decoder $\psi$
				\Repeat
				\State $z_i=\phi(x_i)$, $i\in \{1,\cdots N\}$
				\State assign $\{z_i\}$ to $\{\widehat{z}_i\}$ by Algorithm \ref{algo:least-cost} and obtain $\{S_i\}_{i=1}^m$
				\For{$k=1\to m$}
				\State compute $\nabla_{\theta_{k}} \left(f_k(\theta_{k})\right)$ with $S_k$
				\State sample $S^{(k)}$ of n random points in $\{z_i\}_{i=1}^N$
				\State compute $\nabla_{\theta_{k}} \left(f_{S^{(k)}}(\theta_{k})\right)$ with $S^{(k)}$
				\If{k=1}
				\State $g=\nabla_{\theta_{k}} \left(f_k(\theta_{k})\right)$
				\Else{}
				\State compute $\nabla_{\theta_{k}} \left(f_{S^{(k-1)}}(\theta_{k})\right)$ with $S^{(k-1)}$
				\State $g=\nabla_{\theta_{k}} \left(f_k(\theta_{k})\right)+\alpha\left[\nabla_{\theta_{k}} \left(f_{S^{(k-1)}}(\theta_{k})\right)-\nabla_{\theta_{k-1}} \left(f_{S^{(k-1)}}(\theta_{k-1})\right)\right]$
				\EndIf
				\State $\theta_{k+1}=\theta_{k}-\gamma g$
				\EndFor
				\Until{convergence}
			\end{algorithmic}
		\end{algorithm}
	\end{minipage}
\end{table}		

\section{Theoretical Analysis}
\par From a statistical view, the estimation of discrepancy by the discriminator in GAN is biased and of high variance. Since the discriminator has cumulative preferences of features when classify real and fake data, the estimates of discrepancy are somehow biased. Moreover, as of two-player setting, noise impedes drastically more the training compared to single objective one \citep[][]{chavdarova2019reducing}. Thus, the variance is high. On the contrary, non-adversarial methods treat each data equally and have low variance on estimating the discrepancy. However, since non-adversarial methods are not
accurate enough and not over-parameterized to memorize data, they suffer from errors, which are analysable. Suppose $P_N$ and $Q_N$ are empirical distributions of the sampled data $\{\tilde{z}_i\}_{i=1}^N$ and encoded data $\{z_i\}_{i=1}^N$, while $P_n$ and $Q_n$ denote the empirical distributions of batches with $n$ points sampled from $\{\tilde{z}_i\}_{i=1}^N$ and $\{z_i\}_{i=1}^N$, respectively. We use $\widehat{W}(\cdot,\cdot)$ to denote the estimator of the true Wasserstein distance $W(\cdot,\cdot)$, then the error of estimation can be divided into sampling error $e_s$ and measurement error $e_m$ based on
\begin{equation}
\label{eq:sample and measurment}
\begin{aligned}
\left|\widehat{W}(P_n,Q_n)-W(P_N,Q_N)\right|&\leq \Big|W(P_n,Q_n)-W(P_N,Q_N)\Big|\\
&+\left|\widehat{W}(P_n,Q_n)-W(P_n,Q_n)\right|\\
&= e_s+e_m.
\end{aligned}
\end{equation}
In the following, we elaborate the superiority of the tessellation to reduce $e_s$ and $e_m$ respectively. We also analyze the computational complexity of TWAE in sampling and tessellation procedure.

\subsection{Sampling Error}

\par The target of generative models is to learn a continuous distribution. However, the road to continuity is discrete sampling. Points sampled randomly from the prior distribution are compared with the real data to make the encoder of auto-encoders or generator of GANs smooth in the latent space or the data space, respectively. Thus, while optimizing each batch, the task is to minimize the discrepancy of empirical distributions. Theorem \ref{thm:Gaussian} shows that the sampling error of the GW distance decreases with a rate of $\frac{1}{\sqrt{n}}$, we can also derive similar results for the SW distance.
%
\begin{theorem}
Let $P$, $Q$ is continuous distributions on $\mathbb{R}^d$. Let $P_n$ and $Q_n$ be generated by i.i.d. samples $z_1,\cdots,z_n\sim Q$ and $\tilde{z}_1,\cdots,\tilde{z}_n\sim P$ respectively. $P_n^{'}$ is an independent copy of $P_n$. Let
\begin{equation}
J_2(P)=\int_{w\in S^{d-1}}\int_{F^{-1}_w(0)}^{F^{-1}_w(1)}\frac{F_w(x)\left(1-F_w(x)\right)}{p_w(x)}dxdw,
\end{equation}
where $p_w$ is the density of distribution $P_w$ and $P_w$ is the probability distribution $X^Tw$ where $X\sim P$. $F_w$ is the cumulative function of $P_w$.
If $\max\{J_2(P),J_2(Q)\}<\infty$, then with $n$ approaching infinity
$$
\sqrt{n}\left(SW_2^2\left(P_n,Q_n\right)-SW_2^2\left(P,Q\right)\right)\to N(0,\sigma^2),
$$
$$
nSW_2^2\left(P_n,P_n^{'}\right)\to \gamma_4,
$$
where $\sigma^2$ is the variance correlated with $P$ and $Q$, and $\gamma_4$ is a random variable correlated with $P$.
	\label{thm:sw}
\end{theorem}
\begin{proof}
It is a simple generalization of Theorem \ref{thm: onedim} by integrating on $S^{d-1}$.
\end{proof}

\begin{figure}[hpbt]
	\centering
	\includegraphics[width=0.6\columnwidth]{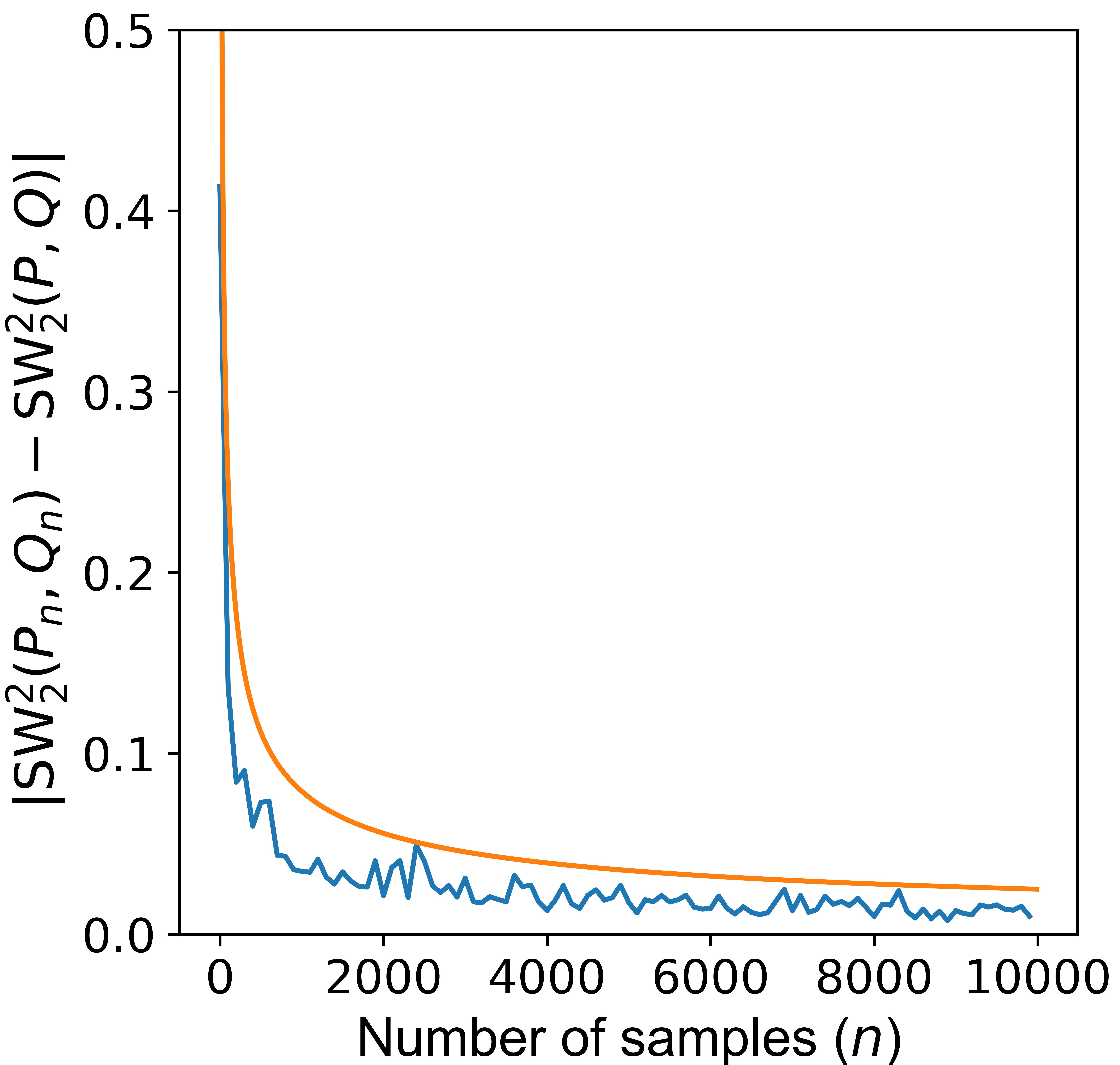}\\
	\includegraphics[width=0.6\columnwidth]{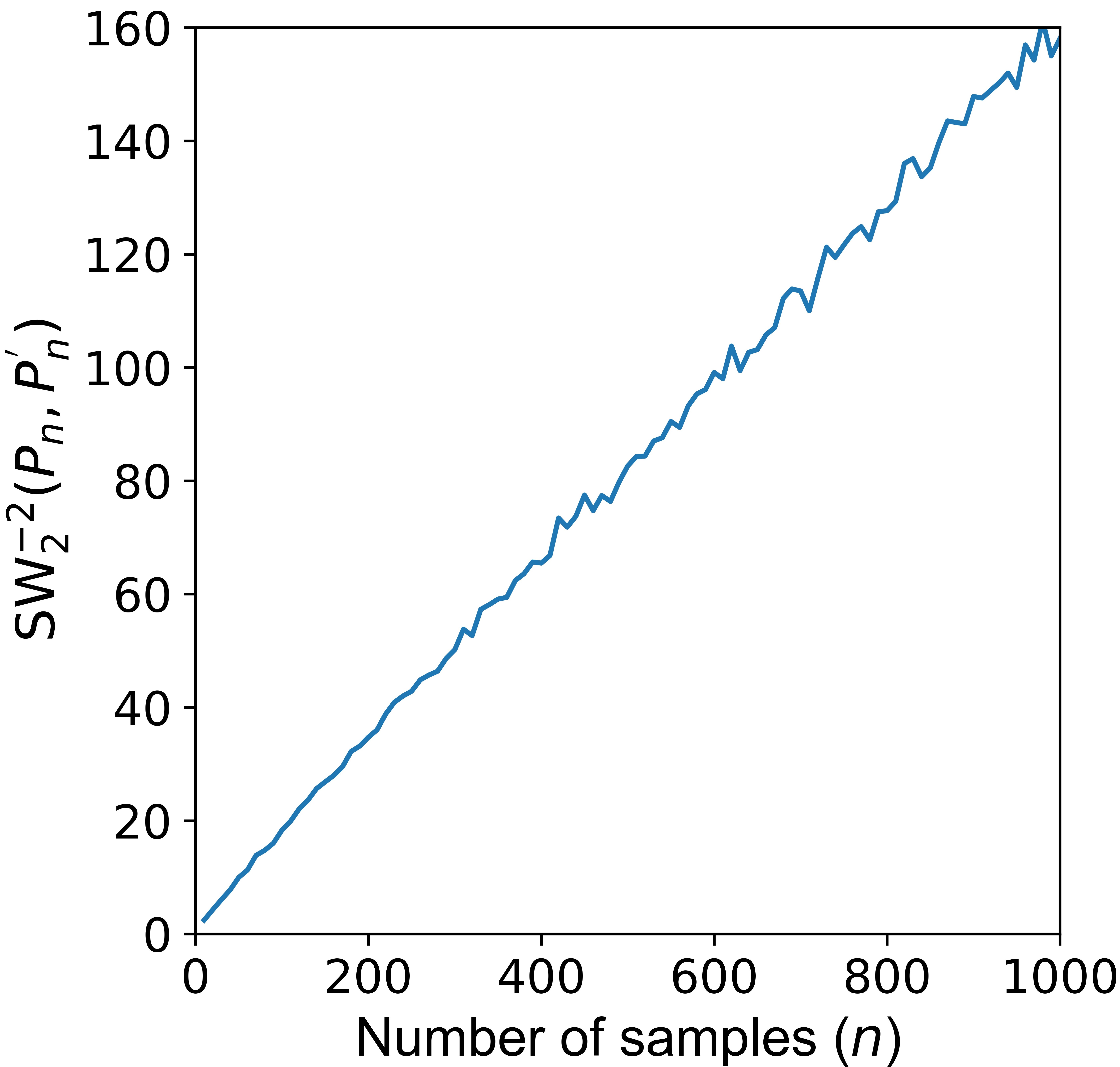}
	\caption{Illustration of the asymptotic property of the sliced-Wasserstein (SW) distance. Here $P_n$ and $P_n^{'}$ are sampled from the same Gaussian distribution $P$ of 64-dimension. $Q_n$ is sampled from a uniform distribution in the unit ball of 64-dimension. $|\operatorname{SW}_2^2(P_n,Q_n)-\operatorname{SW}_2^2(P,Q)|$ is bounded by the orange line of $\frac{C}{\sqrt{n}}$, while the reciprocal of the SW distance $\operatorname{SW}_2^{-2}(P_n,P_n^{'})$ increases linearly with $n$.}
	\label{fig:asymptotic}
\end{figure}

Numerical test simulates the asymptotic property of the SW distance (Fig. \ref{fig:asymptotic}) and we observe that $|SW_2^2(P_n,Q_n)-SW_2^2(P,Q)|$ and $SW_2^2(P_n,P_n^{'})$ decrease roughly via $\mathcal{O}\left(\frac{1}{\sqrt{n}}\right)$ and $\mathcal{O}\left(\frac{1}{n}\right)$, respectively. Then we can obtain upper bounds correlated with $n$, which are tighter than Claim 1 in \citet{deshpande2018generative},
\begin{equation}
\mathbb{E}\left[\left|SW_2^2\left(P_n,Q_n\right)-SW_2^2\left(P,Q\right)\right|\right]\leq \frac{C_1}{\sqrt{n}},
\end{equation}
\begin{equation}
\mathbb{E}\left[SW_2^2\left(P_n,P_n^{\prime}\right)\right]\leq \frac{C_2}{n},
\end{equation}
where $C_1$ and $C_2$ are two constants.

\par At the end of this section, we illustrate the benefit of tessellation intuitively. Fix the number $m$ of optimization step and the batch size $n$ in each step. For $N=mn$, assume that the tessellation procedure induces an extra error which is no larger than $\mathcal{O}\left(\frac{1}{\sqrt{N}}\right)$, i.e.,
\begin{equation}
\frac{1}{m}\sum_{j=1}^{m}W_2^2(P_{n|V_j},Q_{n|S_j})=W_2^2(P_N,Q_N)+\mathcal{O}\left(\frac{1}{\sqrt{N}}\right).
\end{equation}
Then the sampling error induced by the tessellated Wasserstein distance is of the same order with that of $W_2^2(P_N,Q_N)$, i.e., $\mathcal{O}\left(\frac{1}{\sqrt{N}}\right)$. If we sample empirical distributions of $n$ points from $P_N$ and $Q_N$ for $m$ times (denoted by $P_n^{(i)}$ and $Q_n^{(i)}$ for $i$-th time), then the sampling error induced by $\frac{1}{m}\sum_{i=1}^{m}W_2^2(P_n^{(i)},Q_n^{(i)})$ is $\mathcal{O}\left(\frac{1}{\sqrt{n}}\right)$. One thing needs to clarify is that increasing the batch size $n$ can also reduce the sampling error. However, it has drawbacks: 1) larger batch size leads to more consumption on both time and memory; the model optimized with large batch size may converge to saddle points \citep{li2017batch}, which may offset the reduction of sampling error. In other words, if we optimize with batches of size $n$ on $\Omega$, then after a few epochs, $W_2^2(P_n,Q_n)$ is approximately equal to $W_2^2(P_n,P_n^{'})$, where $P_n^{'}$ is an independent copy of $P_n$. This means we cannot identify $Q$ from $P$ with $n$ sampled points. However, if we take a look at a region $V_i$ with probability $P(V_i)=\frac{1}{m}$, we can still find differences between $P_{n|V_i}$ and $Q_{n|V_i}$ because in the past batches only a few points located in $V_i$ and the sampling error was high. So the local information is lost in this way. On the contrary, TWAE samples a batch from each region, so that with the same size of batches, we can approximate the continuous distribution better. Numerical experiments in Section 5 demonstrate the effectiveness of this idea.

\subsection{Measurement Error}
In this section, we illustrate the optimality of using CVT in reducing the measurement error $e_m$ in (\ref{eq:sample and measurment}) and prove the descent rate of $e_m$ with respect to the number of regions $m$ in CVT is $\mathcal{O}(\frac{1}{\sqrt{m}})$.
\par Let $P_n$ and $Q_n$ denote the empirical distribution of $n$ points sampled from the prior distribution and the encoded data set,respectively. Both the SW and GW discrepancy metrics may lead to inaccurate estimation of the discrepancy. For the SW distance, we replace the integration in (\ref{eq:slicedwasserstein}) over $S^{d-1}$ with a summation over a randomly chosen set of unit vectors $\widehat{S}^{d-1}$. For the GW distance, we approximate $P_n$ and $Q_n$ with Gaussian distributions. To reduce the measurement error, we expect that the sum of errors for measuring the discrepancies on the tessellated supports is smaller than that on the whole support. For instance, if we approximate $P_{n|V_j}$ with a Gaussian distribution in each region of $\Omega$, we are actually utilizing a Gaussian mixture model to approximate $P_N$. A standard result in Bayesian nonparametrics says that every probability density is closely approximable by an infinite mixture of Gaussians. However, since the distribution in $\mathbb{R}^d$ is no longer embeddable in the function space $L_p(0,1)$ via quantile functions, it is hard to show the reduction of error with the increase of $m$. The extreme cases make the measurement error hard to analyze theoretically. To exclude the extreme cases, we unify the measurement error induced by different approaches with a parameter $\epsilon$ which depicts the estimator $\widehat{W}$.
\begin{myDef}
	Suppose $P_n$ and $Q_n$ are empirical distributions of $n$ points. An estimator $\widehat{W}$ is $\epsilon$-good for $(P_n,Q_n)$ if it holds that
	\begin{equation}
	\label{eq:egood}
	|\widehat{W}_2^2(P_n,Q_n)-W_2^2(P_n,Q_n)|\leq \epsilon(\operatorname{tr}\left(\Sigma(P_n)\right)+\operatorname{tr}\left(\Sigma(Q_n)\right)),
	\end{equation}
	 where $\Sigma(P_n)$, $\Sigma(Q_n)$ are the unbiased empirical covariance matrices of $P_n$ and $Q_n$ respectively, and $\operatorname{tr}(\cdot)$ is the trace operator.
\end{myDef}

\par To explain the connection of $|\widehat{W}_2^2(P,Q)-W_2^2(P,Q)|$ and $\Sigma(P)$ and $\Sigma(Q)$, for instance, while adopting the GW distance as the estimator $\widehat{W}$, we use multivariate Gaussians to approximate $P$ and $Q$, and ignore the information in the moments higher than two. Intuitively, by doing Taylor expansion on $|\widehat{W}_2^2(P,Q)-W_2^2(P,Q)|$, the loss of moments higher than two can be bounded by the variance of $P$ and $Q$. More formally, we have the following theorem.
\begin{theorem}
	$W_2^2(P_n,Q_n)\leq \frac{2(n-1)}{n-4}\left(\tr\left(\Sigma(P_n)\right)+\tr\left(\Sigma(Q_n)\right)\right)$.
\end{theorem}
\begin{proof}
	First, using the triangle inequality, we have
\begin{equation}
\label{eq:W and variance}
\begin{aligned}
W_2^2(P_n,Q_n) &=\min_{\sigma}\frac{1}{n}\sum_{i=1}^{n}\|z_i-\tilde{z}_{\sigma(i)}\|_2^2\\
&=\min_{\sigma}\frac{1}{n}\sum_{i=1}^{n}\|z_i-\mathbb{E}_{z\sim P_n}[z]+\mathbb{E}_{z\sim P_n}[z]-\mathbb{E}_{z\sim Q_n}[z]+\mathbb{E}_{z\sim Q_n}[z]-\tilde{z}_{\sigma(i)}\|_2^2\\
&\leq\frac{2}{n}\sum_{i=1}^{n}\|z_i-\mathbb{E}_{z\sim P_n}[z]\|_2^2+\frac{2}{n}\sum_{i=1}^{n}\|\tilde{z}_i-\mathbb{E}_{z\sim Q_n}[z]\|_2^2+2\|\mathbb{E}_{z\sim P_n}[z]-\mathbb{E}_{z\sim Q_n}[z]\|_2^2\\
&=\frac{2(n-1)}{n}\left(\tr\left(\Sigma\left(P_n\right)\right)+\tr\left(\Sigma\left(Q_n\right)\right)\right)+2\|\mathbb{E}_{z\sim P_n}[z]-\mathbb{E}_{z\sim Q_n}[z]\|_2^2.
\end{aligned}
\end{equation}
Note that
\begin{equation}
\label{eq:mean and W}
\begin{aligned}
\|\mathbb{E}_{z\sim P_n}[z]-\mathbb{E}_{z\sim Q_n}[z]\|_2^2&=\|\frac{\sum_{i=1}^{n}z_i}{n}-\frac{\sum_{i=1}^{n}\tilde{z}_i}{n}\|_2^2\\
&=\frac{1}{n^2}\|\sum_{i=1}^{n}\left(z_i-\tilde{z}_{\sigma(i)}\right)\|_2^2\\
&\leq\frac{2}{n^2}\min_{\sigma}\sum_{i=1}^{n}\|z_i-\tilde{z}_{\sigma(i)}\|_2^2\\
&=\frac{2}{n}W^2_2(P_n,Q_n).
\end{aligned}
\end{equation}
By taking (\ref{eq:mean and W}) into (\ref{eq:W and variance}), we have
\begin{equation}
W_2^2(P_n,Q_n)\leq \frac{2(n-1)}{n-4}\left(\tr\left(\Sigma(P_n)\right)+\tr\left(\Sigma(Q_n)\right)\right).
\end{equation}
\end{proof}

In general, if the estimator is good, then the measurement error $|\widehat{W}_2^2(P,Q)-W_2^2(P,Q)|$ should be a high-level minim to the upper bound of $W_2^2(P,Q)$. Thus it is natural to assume $\widehat{W}_2$ satisfying (\ref{eq:egood}) with a relatively small $\epsilon$. With the assumption that $\widehat{W}_2$ is $\epsilon$-good, we derive the optimality of using CVT in the setting of TWAE.
\begin{theorem}\label{thm:twd}
	Let $P$ be a uniform distribution on a compact and connected set $\Omega$ and $Q$ is the target distribution. Assume that the optimal transport map $\mathcal{T}$ from $P$ to $Q$ is continuously differentiable. Let $\{V_i\}_{i=1}^m$ be a tessellation on $\Omega$. Assume that the estimator $\widehat{W}_2$ is $\epsilon$-good for $\{(P_{n|V_i},\mathcal{T}_{\#}P_{n|V_i})\}_{i=1}^m$.\\
	\\
	1. The expectation of measurement error of the tessellated Wasserstein distance is upper bounded, i.e.,
	\begin{equation}
	\label{eq: inequl}
	\mathbb{E}[e_m]\leq C_3\sum_{i=1}^{m}\int_{V_i}\|z-\hat{z}_i\|_2^2dz,
	\end{equation}
	where $\widehat{z}_i$ is the mass centroid of $V_i$ and $C_3$ is a constant correlated with $\Omega$, $\epsilon$ and $Q$. \\
	\\
	2. A necessary condition for the right side of (\ref{eq: inequl}) to be minimized is that $\{V_i\}_{i=1}^m$ is the CVT and $\{\widehat{z}_i\}_{i=1}^m$ is the generator set.

\end{theorem}

\begin{proof}
	The measurement error of the tessellated Wasserstein distance can be formulated as
	\begin{equation}
	e_m=\sum_{i=1}^{m}P(V_i)\left|W_2^2\left(P_{n|V_i},\mathcal{T}_{\#}P_{n|V_i}\right)-\widehat{W}_2^2\left(P_{n|V_i},\mathcal{T}_{\#}P_{n|V_i}\right)\right|.
	\end{equation}
	Since the estimator $\widehat{W}_2$ is $\epsilon$-good, we have
	\begin{equation}
	\begin{aligned}
	e_m&\leqslant \epsilon \sum_{i=1}^{m} P\left(V_{i}\right)\left(\operatorname{tr}\left(\Sigma\left(\mathcal{T}_{\#}P_{n|V_i}\right)\right)+\operatorname{tr}\left(\Sigma\left(P_{n | V_{i}}\right)\right)\right) \\
	&= \frac{\epsilon n}{n-1} \sum_{i=1}^{m} P\left(V_{i}\right) \mathbb{E}_{z\sim P_{n|V_i}}\left[\|\mathcal{T}(z)-\bar{\mathcal{T}}_i\|_{2}^{2}+\|z-\bar{z}_i\|_{2}^{2}\right], \\
	\label{eq:proof}
	\end{aligned}
	\end{equation}
	where $\bar{z}_i=\mathbb{E}_{z\sim P_{n|V_i}}\left[z\right]$, $\bar{\mathcal{T}}_i=\mathbb{E}_{z\sim P_{n|V_i}}\left[\mathcal{T}(z)\right]$. Note that $\mathcal{T}$ is continuously differentiable on the compact set $\Omega$, then $\mathcal{T}$ is Lipschitz continuous with a constant $L$. Thus
	\begin{equation}
	\label{eq: lipschitz}
	\mathbb{E}_{z\sim P_{n|V_i}}\left[\|\mathcal{T}(z)-\bar{\mathcal{T}}_i\|_{2}^{2}\right]\leq L^2 \mathbb{E}_{z\sim P_{n|V_i}}\left[\|z-\bar{z}_i\|_{2}^{2}\right].
	\end{equation}
	By taking (\ref{eq: lipschitz}) into (\ref{eq:proof}), we obtain
	\begin{equation}
	\label{eq: lipschitz2}
	\begin{aligned}
	e_m&\leqslant \frac{\epsilon n}{n-1} \sum_{i=1}^{m} P\left(V_{i}\right)\left(1+L^{2}\right) \mathbb{E}_{z\sim P_{n|V_i}}\left[\|z-\bar{z}_i\|_{2}^{2}\right]  \\
	&=\frac{\epsilon n\left(1+L^{2}\right)}{(n-1)|\Omega|} \sum_{i=1}^{m} |V_i| \mathbb{E}_{z\sim P_{n|V_i}}\left[\|z-\bar{z}_i\|_{2}^{2}\right].
	\end{aligned}
	\end{equation}
	 The last equality is because $P$ is a uniform distribution on $\Omega$, thus $P(V_i)=\frac{|V_i|}{|\Omega|}$. Let $\tilde{z}_1^{(i)},\cdots,\tilde{z}_n^{(i)}$ be the support points of $P_{n|V_i}$, since they are randomly sampled from the uniform distribution on $V_i$, then
	 \begin{equation}
	 \label{eq: expectation}
	 \begin{aligned}
	 \mathbb{E}_{P_{n|V_i}}\left[\mathbb{E}_{z\sim P_{n|V_i}}\left[\|z-\bar{z}_i\|_{2}^{2}\right]\right]&=\mathbb{E}_{P_{n|V_i}}\left[\frac{1}{n}\sum_{k=1}^{n}\|\tilde{z}_k^{(i)}-\bar{z}^{(i)}\|_2^2\right]\\
	 &= \frac{n-1}{n|V_i|}\int_{V_i}\|z-\widehat{z}_i\|_2^2 d z,
	 \end{aligned}
	 \end{equation}
	 where $\widehat{z}_i=\frac{1}{|V_i|}\int_{V_i}z dz$. Thus combining (\ref{eq: expectation}) and (\ref{eq: lipschitz2}), we have
	 \begin{equation}
	 \begin{aligned}
	 \mathbb{E}[e_m]&\leqslant \frac{\epsilon n\left(1+L^{2}\right)}{(n-1)|\Omega|} \sum_{i=1}^{m} |V_i|\mathbb{E}_{P_{n|V_i}} \left[\mathbb{E}_{z\sim P_{n|V_i}}\left[\|z-\bar{z}_i\|_{2}^{2}\right]\right].  \\
	 &=\frac{\epsilon\left(1+L^2\right)}{|\Omega|}\sum_{i=1}^{m}\int_{V_i}\|z-\hat{z}_i\|_2^2dz.
	 \end{aligned}
	 \end{equation}
	 Let $C_3=\frac{\epsilon\left(1+L^2\right)}{|\Omega|}$, then we obtain the inequality in (\ref{eq: inequl}).
	
	Next, we prove CVT is  the necessary condition to minimize the upper bound in (\ref{eq: inequl}). First, fix the tessellation $\{V_i\}_{i=1}^m$, $\forall j \in \{1,\cdots,m\}$
	\begin{equation}
	\begin{aligned}
	\int_{V_j}\|z-{z}_j^{*}\|_2^2 d z= \int_{V_j}\left(\|z\|_2^2-2z^T{z}_j^{*}+\|z_j^{*}\|_2^2\right) dz.
	\end{aligned}
	\end{equation}
	The integration is minimized when $z_{j}^{*}=\widehat{z}_j=\frac{1}{|V_j|}\int_{V_j}z dz$. Second, fix $\widehat{z}$ and see what happens if $\{V_i\}_{i=1}^m$ is not a Voronoi tessellation generated by $\widehat{z}$. Suppose that $\{\widehat{V}_i\}_{i=1}^m$ is the Voronoi tessellation generated by $\widehat{z}$. Since $\{V_i\}_{i=1}^m$ is not a Voronoi tessellation, there exists a particular value of $z\in V_i$, $\exists j\in\{1,\cdots,m\}$ that
	\begin{equation}
	\|z-\widehat{z}_j\|_2^2<\|z-\widehat{z}_i\|_2^2.
	\end{equation}
	Thus,
	\begin{equation}
	\sum_{i=1}^{m} \int_{V_i}\|z-\widehat{z}_i\|_2^2 d z>\sum_{i=1}^{m} \int_{\widehat{V}_i}\|z-\widehat{z}_i\|_2^2 d z.
	\end{equation}
	So that the upper bound is minimized when $\{V_i\}_{i=1}^m$ is chosen to be the CVT and $\widehat{z}$ is the set of generators.
\end{proof}
\begin{theorem}\label{thm:decrease}
	In the setting of Theorem \ref{thm:twd}, let $\{V^{*}_i\}_{i=1}^m$ and $\{\widehat{z}^{*}_i\}_{i=1}^m$ be a tessellation on $\Omega$ and its generator set which minimize the right side of inequality (\ref{eq: inequl}), i.e.,
	\begin{equation}\label{eq:assumption}
\{V^{*}_i\}_{i=1}^m,\{\widehat{z}^{*}_i\}_{i=1}^m\in \arg\min_{\{V_i\}_{i=1}^m,\{\widehat{z}_i\}_{i=1}^m} C_3\sum_{i=1}^{m}\int_{V_i}\|z-\hat{z}_i\|_2^2dz,
	\end{equation}
	then the expectation of $e_m$ with respect to $\{V^{*}_i\}_{i=1}^m$ holds that
	\begin{equation}
	\mathbb{E}[e_m]\leq\frac{C_4}{\sqrt{m}},
	\end{equation}
	where $C_4$ is a constant correlated with $\Omega$, $\epsilon$ and $Q$.
\end{theorem}

\begin{proof}
	Following the result in Theorem \ref{thm:twd}, we have
	\begin{equation}
	C_3 \sum_{i=1}^{m} \int_{V^{*}_i}\|z-\widehat{z}^{*}_i\|_2^2 d z
	=C_3 \sum_{i=1}^{m}|V^{*}_i|\mathbb{E}_{P_{n|V^{*}_i}}\left[\mathbb{E}_{z\sim P_{n|V^{*}_i}}\left[\|z-\widehat{z}^{*}_i\|_{2}^{2}\right]\right].
\label{eq:thm4}
	\end{equation}
	Since (\ref{eq:assumption}) holds, following the result in Theorem \ref{thm:twd}, $\{V^{*}_i\}_{i=1}^m$ is a CVT and $\{\widehat{z}^{*}_i\}_{i=1}^m$ is its generator. Let
	\begin{equation}
	\begin{aligned}
	P^{*}&=\sum_{i=1}^{m}\frac{|V^{*}_i|}{|\Omega|}P_{n|V^{*}_i},\\
	Q^{*}&=\sum_{i=1}^{m}\frac{|V^{*}_i|}{|\Omega|}\delta_{\widehat{z}^{*}_i}.
	\end{aligned}
	\end{equation}
 Suppose $\mathcal{T}_1$ is the optimal transport map from $P^{*}$ to $Q^{*}$, then let $z$ belong to the support of$P_{n|V^{*}_i}$, $\mathcal{T}_1(z)=\widehat{z}_i^{*}$, which is held for $i=1,\cdots,m$. Thus, we have
	\begin{equation}
	\sum_{i=1}^{m}\frac{|V^{*}_i|}{|\Omega|}\mathbb{E}_{P_{n|V^{*}_i}}\left[\mathbb{E}_{z\sim P_{n|V^{*}_i}}\left[\|z-\widehat{z}^{*}_i\|_{2}^{2}\right]\right]=\mathbb{E}_{P^{*}}\left[W_2^2(P^{*},Q^{*})\right].
	\label{eq:m}
	\end{equation}
	Since $P^{*}$ is an empirical distribution, let $P^{*}_m$ be an empirical distribution of $m$ points i.i.d. sampled from $P^{*}$. Since (\ref{eq:assumption}) holds, we have
	\begin{equation}
	\begin{aligned}
	C_3 \sum_{i=1}^{m} \int_{V^{*}_i}\|z-\widehat{z}^{*}_i\|_2^2 d z
	&=C_3|\Omega|\mathbb{E}_{P^{*}}\left[W_2^2(P^{*},Q^{*})\right]\\
	&\leq C_3|\Omega|\mathbb{E}_{P^{*}}\left[\mathbb{E}_{P^{*}_m}\left[W_2^2(P^{*},P_m^{*})\right]\right].\\
	\end{aligned}
	\end{equation}
	For fixed $P^{*}$, according to Theorem \ref{thm:sommer}, with $m$ approaching infinity, $\sqrt{m}W_2^2(P^{*},P^{*}_m)$ converges to a distribution. Since the support sets of $P^{*}$ and $P^{*}_m$ belong to $\Omega$ which is compact, then there exists a constant $C_4^{\prime}$ such that
	\begin{equation}
	\mathbb{E}_{P^{*}}\left[\mathbb{E}_{P^{*}_m}\left[W_2^2(P^{*},P_m^{*})\right]\right]\leq \frac{C_4^{\prime}}{\sqrt{m}}.
	\end{equation}
	Let $C_4=\frac{C^{\prime}_{4}}{C_3|\Omega|}$, which is correlated with $\Omega$, $\epsilon$ and $Q$. Finally, we obtain
	\begin{equation}
	\mathbb{E}[e_m]\leq C_3 \sum_{i=1}^{m} \int_{V^{*}_i}\|z-\widehat{z}^{*}_i\|_2^2 d z \leq \frac{C_4}{\sqrt{m}},
	\end{equation}
	which completes the proof.
\end{proof}

Since in the training procedure we need to define the tessellation $\{V_i\}_{i=1}^m$ before $Q$ is known, the upper bound of error corresponding to $\{V_i\}_{i=1}^m$ is of importance. Theorem \ref{thm:twd} gives the reason for utilizing the CVT technique and Theorem \ref{thm:decrease} shows that the error decreases with a rate of $\frac{1}{\sqrt{m}}$. Note that after a few iterations, $Q$ is approximately equal to $P$, then the optimal transport map $\mathcal{T}$ is almost identical. Thus, $\mathcal{T}(V_i)\approx V_i$ is a set of points that are closest to $\widehat{z}_i$ other than $\widehat{z}_j(j\neq i)$. So the empirical distribution of $S_i$ obtained by (\ref{eq:hitchcock}) is close to $\mathcal{T}_{\#}P_{n|V_i}$. Thus, in the algorithm, we compute $W(P_{n|V_i},Q_{n|S_i})$ instead of $W(P_{n|V_i},\mathcal{T}_{\#}P_{n|V_i})$. If $\{V_i\}_{i=1}^m$ is not a CVT, $\mathcal{T}_{\#}P_{n|V_i}$ and $Q_{n|S_i}$ will not coincide. The error induced by the approximation of $Q_{n|S_i}$ to $\mathcal{T}_{\#}P_{n|V_i}$ is hard to model. Nevertheless, it makes little effect on the results in the experiment.
\subsection{Sampling and tessellation complexity}
In TWAE, for an arbitrary $z$ sampled from the uniform distribution on the unit ball in the latent space, we need to justify which region it belongs to by computing the distance between $z$ and the generators of each region. The complexity of this is of order $\mathcal{O}(Nm)$ . The tessellation complexity mainly arises in sorting the elements of the distance matrix $M$ , which is $\mathcal{O}(Nm\log(Nm))$ . Since we only sort the element of $M$ once for $N$ data points, the tessellation complexity $\mathcal{O}(Nm\log(Nm))$. By contrast, the sampling complexity of WAE is $\mathcal{O}(N)$ . Though the complexity of TWAE is higher than that of WAE, the time consumption of LCM on CPU is comparable to that of the backpropagation algorithm on GPU (Table 1).

\section{Experimental Results}
\par In this section, we numerically evaluate TWAE from five aspects. In section 5.3, we compare TWAE with related studies. In section 5.4, we test the optimization method introduced in section 3. In section 5.5, we compare the performance of the CVT technique and sphere packing. In section 5.6, we compare the models with and without tessellation. We test TWAE with the GW distance (TWAE-GW) and the SW distance (TWAE-SW) respectively on two real-world datasets including MNIST \citep{lecun1998gradient} consisting of 70k images and CelebA \citep{liu2015deep} consisting of about 203k images. Finally, in section 5.7, we test TWAE with other SW distances such as DSW, Max-SW and GSW distances on LSUN-Bedrooms dataset \citep{yu2015lsun}. We use the Fr\'{e}chet inception distance (FID) introduced by \citet{heusel2017gans} to measure the quality of the generated images. Smaller FID indicates better quality.

\subsection{Architectures for different datasets}
For MNIST, we use a simple auto-encoder consisting of a mirrored deep convolutional neural network with ReLu as the activation function to compare the performance of the CVT technique and sphere packing (Section 5.5).
\par Encoder architecture:
\begin{equation*}
\begin{aligned}
x \in \mathcal{R}^{28 \times 28} & \rightarrow \text {Conv}_{128} \rightarrow \mathrm{BN} \rightarrow \mathrm{ReLU} \\
& \rightarrow \mathrm{Conv}_{256} \rightarrow \mathrm{BN} \rightarrow \mathrm{ReLU} \\
& \rightarrow \mathrm{Conv}_{512} \rightarrow \mathrm{BN} \rightarrow \mathrm{ReLU} \\
& \rightarrow \mathrm{Conv}_{1024} \rightarrow \mathrm{BN} \rightarrow \mathrm{ReLU} \rightarrow \mathrm{FC}_{8}.
\end{aligned}
\end{equation*}
\par Decoder architecture:
\begin{equation*}
\begin{aligned}
z \in \mathcal{R}^{8} & \rightarrow \mathrm{FC}_{7 \times 7 \times 1024} \\
& \rightarrow \mathrm{FSConv}_{512} \rightarrow \mathrm{BN} \rightarrow \mathrm{ReLU} \\
& \rightarrow \mathrm{FSConv}_{256} \rightarrow \mathrm{BN} \rightarrow \mathrm{ReLU} \rightarrow \mathrm{FSConv}_{1}.
\end{aligned}
\end{equation*}

\par
\begin{figure}[t!]
	\centering
	\includegraphics[width=0.70\columnwidth]{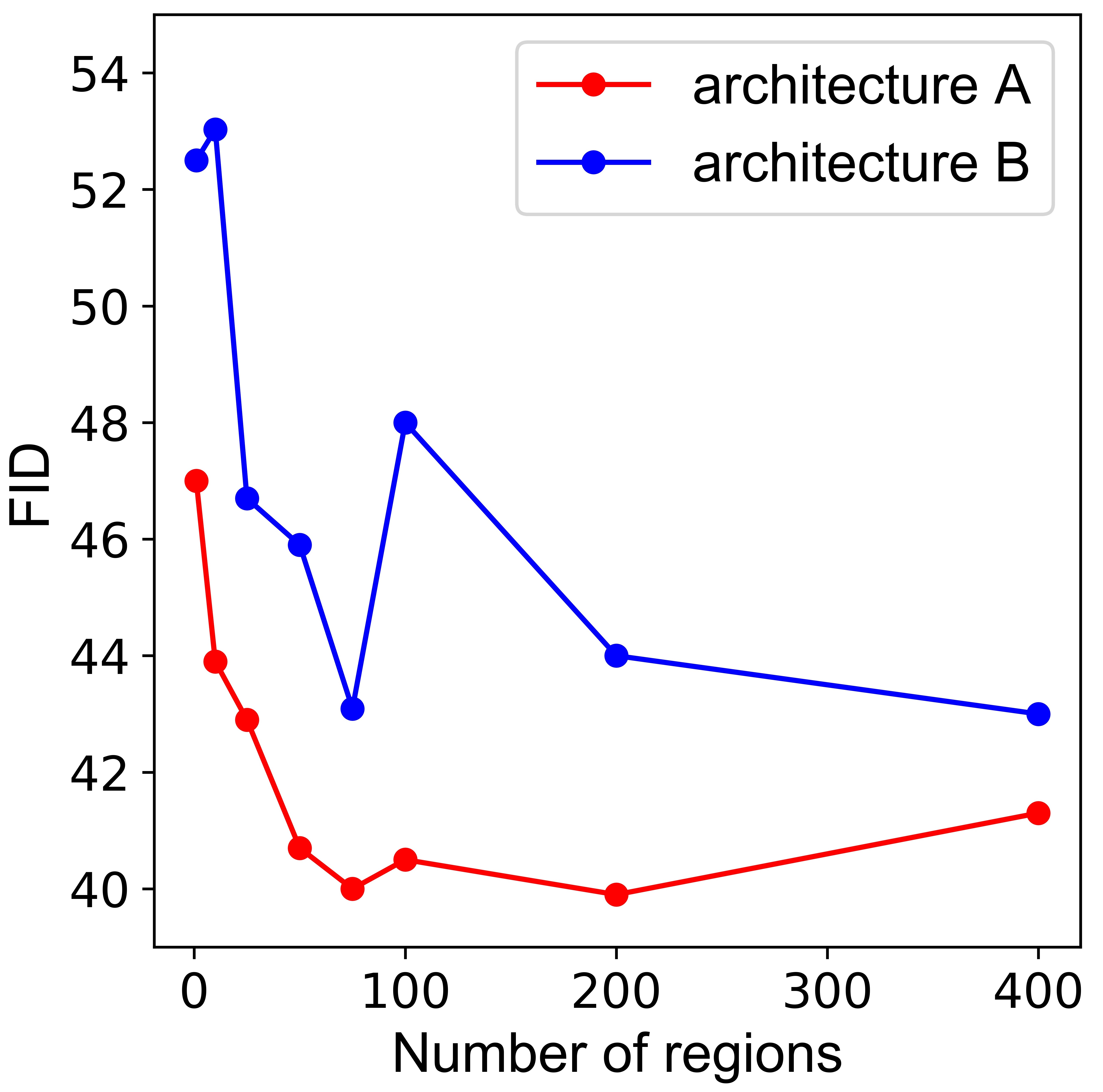}
	\caption{Comparison of changing trend of FID scores versus the number of regions $m$. Here $m$ are set to 1 (without tessellation), 10, 25, 50, 75, 100, 200, 400 for both architectures A and B. }
	\label{fig:two_acrchi}
\end{figure}

 For CelebA, we use two architectures A and B with different sizes of parameters to test if TWAE shows consistent results under different architectures (Fig. \ref{fig:two_acrchi}). Numerical experiments show that our tessellation technique is effective on both architectures. The FID score decreases rapidly when the number of regions $m$ is lower than 100. However, there is no more decline when $m$ is larger. Architecture A is similar to that of \citet{tolstikhin2017wasserstein} and is used to compare the performance of TWAE with other generative auto-encoders fairly (Section 5.3).

\par Encoder of architecture A:
\begin{equation*}
\begin{aligned}
x \in \mathcal{R}^{64 \times 64 \times 3} & \rightarrow \operatorname{Conv}_{128} \rightarrow \mathrm{BN} \rightarrow \operatorname{ReLU} \\
& \rightarrow \mathrm{Conv}_{256} \rightarrow \mathrm{BN} \rightarrow \operatorname{ReLU} \\
& \rightarrow \mathrm{Conv}_{512} \rightarrow \mathrm{BN} \rightarrow \mathrm{ReLU} \\
& \rightarrow \mathrm{Conv}_{1024} \rightarrow \mathrm{BN} \rightarrow \mathrm{ReLU} \rightarrow \mathrm{FC}_{64}.
\end{aligned}
\end{equation*}

\par Decoder of architecture A:
\begin{equation*}
\begin{aligned}
z \in \mathcal{R}^{64} & \rightarrow \mathrm{FC}_{8 \times 8 \times 1024} \\
& \rightarrow \mathrm{FSConv}_{512} \rightarrow \mathrm{BN} \rightarrow \mathrm{ReLU} \\
& \rightarrow \mathrm{FSConv}_{256} \rightarrow \mathrm{BN} \rightarrow \mathrm{ReLU} \\
& \rightarrow \mathrm{FSConv}_{128} \rightarrow \mathrm{BN} \rightarrow \mathrm{ReLU} \rightarrow \mathrm{FSConv}_{3}.
\end{aligned}
\end{equation*}

\par Architecture B has the same number of layers and half the number of nodes. For less computational cost, we use architecture B to investigate the properties of TWAE extensively (Sections 5.4 and 5.6). For LSUN-Bedrooms, we use archtecture A since the size of this dataset is much larger than those of CelebA and the MNIST (Section 5.7).
\par Encoder of architecture B:
\begin{equation*}
\begin{aligned}
x \in \mathcal{R}^{64 \times 64 \times 3} & \rightarrow \operatorname{Conv}_{64} \rightarrow \mathrm{BN} \rightarrow \operatorname{ReLU} \\
& \rightarrow \mathrm{Conv}_{128} \rightarrow \mathrm{BN} \rightarrow \operatorname{ReLU} \\
& \rightarrow \mathrm{Conv}_{256} \rightarrow \mathrm{BN} \rightarrow \mathrm{ReLU} \\
& \rightarrow \mathrm{Conv}_{512} \rightarrow \mathrm{BN} \rightarrow \mathrm{ReLU} \rightarrow \mathrm{Conv}_{64}.
\end{aligned}
\end{equation*}
\par Decoder of architecture B:
\begin{equation*}
\begin{aligned}
z \in \mathcal{R}^{64} & \rightarrow \mathrm{FSConv}_{512} \rightarrow \mathrm{BN} \rightarrow \mathrm{ReLU} \\
& \rightarrow \mathrm{FSConv}_{256} \rightarrow \mathrm{BN} \rightarrow \mathrm{ReLU} \\
& \rightarrow \mathrm{FSConv}_{128} \rightarrow \mathrm{BN} \rightarrow \mathrm{ReLU} \\
& \rightarrow \mathrm{FSConv}_{64} \rightarrow \mathrm{BN} \rightarrow \mathrm{ReLU} \rightarrow \mathrm{FSConv}_{3}.
\end{aligned}
\end{equation*}

\subsection{Experimental setup}
The hyperparameter $\lambda$ of the auto-encoder in (\ref{eq:autoencoder}) is set to 1 for SW distance and 0.01 for GW, GSW and DSW distance. The dimensionalities of the latent space are set to 8 for MNIST, 64 for CelebA and 128 for LSUN-Bedrooms, respectively. The number 241 of root lattices of $E_8$-lattice is chosen for sphere packing test.
How many data points ($N$) in the training dataset should be used for one single tessellation is a question. In the traditional setting, the data is shuffled in each epoch to prevent overfitting. If we take $N$ as large as the size of the training dataset, the designed batches in each epoch will be approximately the same, which leads to bad generalization. Thus, larger $N$ may not perform better. We tried various values of $N$ and noticed that $N=10000$ or $20000$ work well. Compared with traditional algorithms, the only extra computation is using LCM to solve the Hitchcock problem to design batches for each data. The time cost of LCM on CPU is comparable to that of backpropagation algorithm (BP) on GPU (Table 1). We implement our algorithms on Pytorch with the Adam optimizer.
\begin{table}[t!]
	
	\centering
	\begin{tabular}{lccccc}
		\toprule
		&\multicolumn{3}{c}{LCM}&\multicolumn{2}{c}{BP}\\
		\cmidrule(lr){2-4}\cmidrule(lr){5-6}
		&$m=100$&$m=200$&$m=400$& Architecture A & Architecture B\\
		\midrule
		$N=4000$&1.43 (0.02)&2.87 (0.01)&7.11 (0.07)& 15.67 (0.08)& 3.35 (0.01)\\
		\midrule
		$N=10000$&7.71 (0.02)&24.34 (0.04)&50.67 (0.16)& 39.30 (0.11)& 6.75 (0.02)\\
		\midrule
		$N=20000$&47.06 (0.01) &96.56 (1.25)&198.07 (1.27)& 78.34 (1.42)  &13.49 (0.03) \\
		\bottomrule
	\end{tabular}\label{table:time_cost}
\caption{Comparison of time cost between LCM on CPU and BP on GPU (seconds)}
\end{table}

\subsection{TWAE can generate high-quality images}
\begin{figure}[hpbt]
	\centering
	\includegraphics[width=0.90\columnwidth]{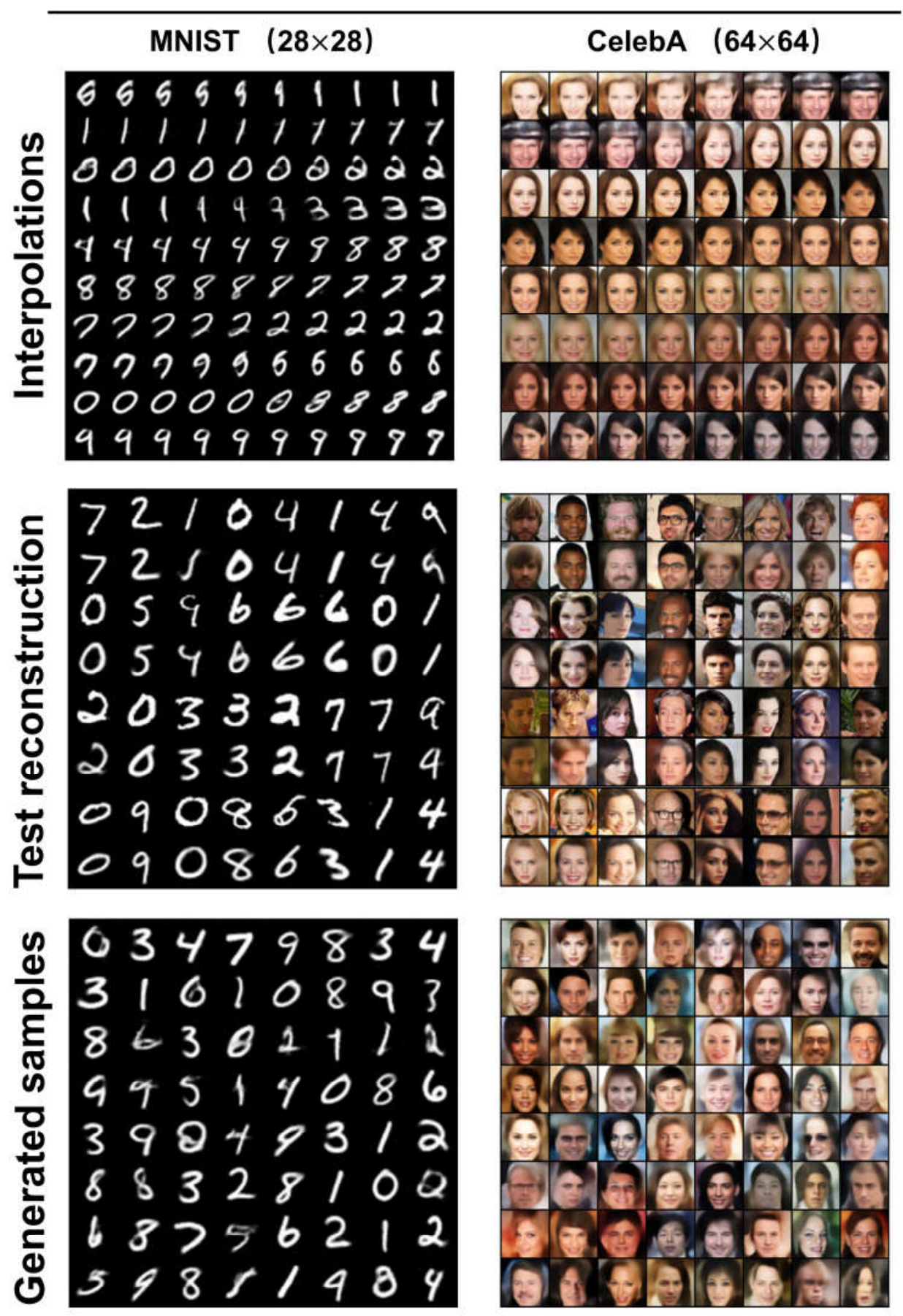}
	\caption{Comparison of interpolated, reconstructed and generated images by TWAE. }
	\label{fig:twae}
\end{figure}

\par We first test if TWAE can approximate the support of the distribution of real data with a smooth and well-learned manifold by interpolations, test reconstruction, and random generating (Fig. \ref{fig:twae}).
For interpolation, considering the probability concentrated near the surface of the unit ball, we interpolate on the curve near the surface instead of linear interpolation to avoid interpolating near the origin. In our experiments, the transition of the decoder from one point to another in the latent space is smooth and gradual. For reconstruction, TWAE can reconstruct the test data which means the model generalizes well. For random generating, samples are generated by sampling in the unit ball uniformly and transforming the resulting vector $z$ into an image via the decoder. By generating images of good quality, the ``hole" in the latent space is filled and TWAE indeed generate a well-learned manifold. We also compared the performance of TWAE with WAE-GAN, WAE-MMD \citep{tolstikhin2017wasserstein}, SWAE and VAE. Only WAE-GAN has a discriminator. We use the results in \citet{tolstikhin2017wasserstein,kolouri2018sliced} since the architectures of these networks are similar, and it is not easy to reproduce the results of WAE-GAN. TWAE shows very competitive performance compared to WAE-GAN (Table 2).
\begin{table}[hpbt]
	\centering
	
	\setlength{\tabcolsep}{7mm}{
		\begin{tabular}{lc}
			\toprule
			\textbf{Model} & \textbf{FID}\\
			\midrule
			TWAE-SW &39.9  \\
			TWAE-GW &44.5    \\
			VAE     &63   \\
			WAE-MMD &55   \\
			SWAE    &79   \\
			\midrule
			WAE-GAN &42   \\
			\bottomrule
		\end{tabular}}\label{table:comparison}
		\caption{Performance comparison of different methods on CelebA}
\end{table}

\subsection{The non-identical batch optimization is effective}
\par We set three different numbers of regions (i.e., $m=$ 100, 200, 400) for both MNIST and CelebA. Numerical results show that the FID score decreases with larger $m$ on MNIST, while it doesn't change significantly on CelebA (Table 3). The difference is probably due to the diverse complexity of the two datasets. We note that the distribution of each batch is different as we put similar data into a batch. The discrepancy of different batches is larger with relatively smaller batch sizes. To address this issue, we propose the non-identical batch optimization method (Section 3) by adding a regularizer for better generalization. Here the hyperparameter $\alpha$ is set to $0.2$. Numerical results of TWAE with the regularizer indeed show better performance than without it with different $m$s for most cases (Table 3). The only exception is that for the GW distance with $m=200$ and 400 respectively, which is explained in Section 5.7.
\begin{table*}[h!]
	
	\centering
	\label{tab:number}
	\begin{tabular}{lcccc}
		\toprule
		& \multicolumn{4}{c}{MNIST}  \\
		\cmidrule(lr){2-5}
		&TWAE-SW&TWAE-SW(r)&TWAE-GW&TWAE-GW(r)\\
		\midrule
		$m=100$&20.4&16.3&18.0&15.9  \\
		$m=200$&17.5&16.0&15.7&14.3  \\
		$m=400$&15.6&13.9  &14.2 &13.8\\
		\bottomrule
	\end{tabular}
	\begin{tabular}{lcccc}
		\toprule
		& \multicolumn{4}{c}{CelebA}  \\
		\cmidrule(lr){2-5}
		&TWAE-SW&TWAE-SW(r)&TWAE-GW&TWAE-GW(r)\\
		\midrule
		$m=100$ &49.2&47.8&46.7&44.5\\
		$m=200$ &50.2&44.1&47.2&48.1\\
		$m=400$ & 47.2  &43.5&54.0&57.2\\
		\bottomrule
	\end{tabular}
\caption{Performance comparison of TWAE with or without regularizer on MNIST and CelebA with three given numbers of regions}
\end{table*}

\subsection{The CVT technique gets similar performance with the exact model}

\par
\begin{figure}[t!]
	\centering
	\includegraphics[width=\columnwidth]{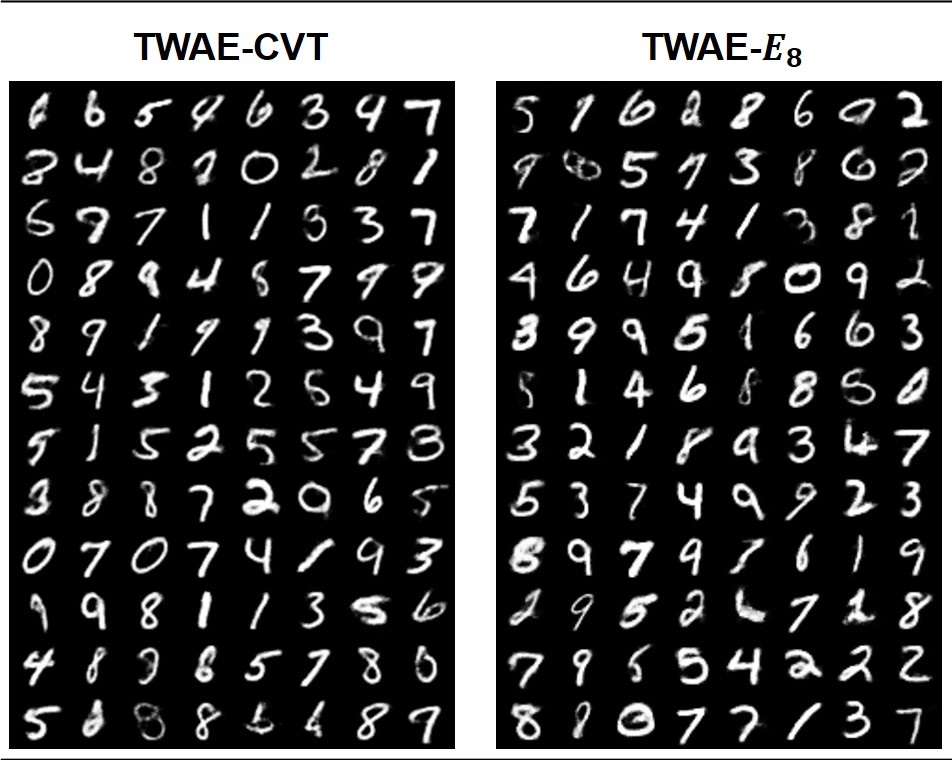}
	\caption{Comparison of generated images of TWAE with CVT and $E_8$-lattice. The FID scores of them are 16.8 and 16.9 respectively.}
	\label{fig:lattice}
\end{figure}

 The CVT technique is an iterative and approximate algorithm that can be adjusted to any dimensions. The iteration is based on integrating over each region. The computation goes up exponentially as the dimension increases. So the CVT technique may not be accurate enough in high-dimensional cases. Thus, it is necessary to explore the effect of it. We implement TWAE with exact lattices and compare its performance with that of the CVT technique. For the MNIST dataset, the dimension of the latent space is 8. Numerical results show that the CVT technique achieves comparable performance and gets very similar FID score with the sphere packing $E_8$-lattice dividing into 241 regions (Fig. \ref{fig:lattice}), indicating that it gets similar performance with the exact model.

\subsection{Tessellation indeed improve the performance of generation}

\par
\begin{figure}[t!]
	\centering
	\includegraphics[width=1.00\columnwidth]{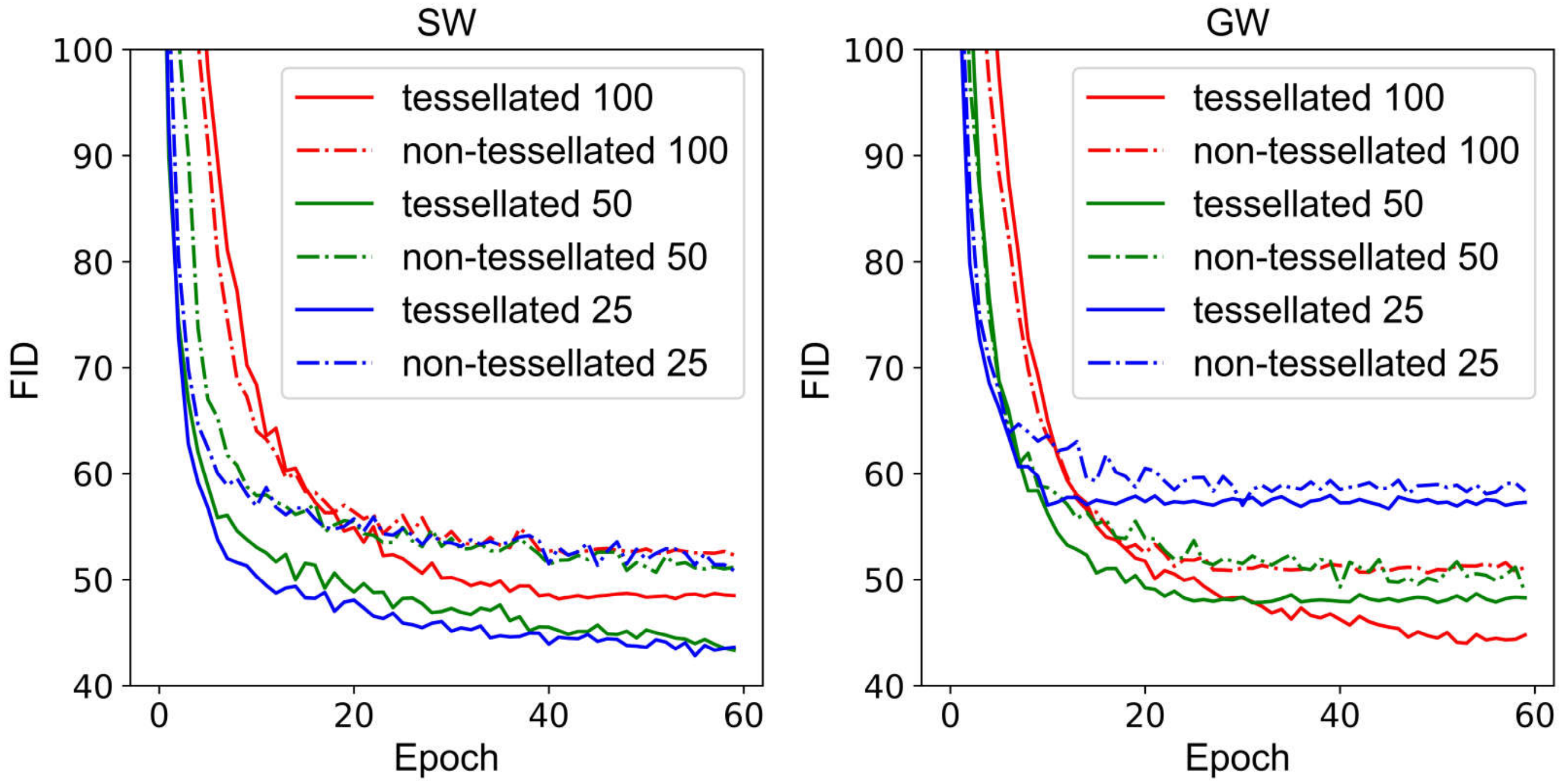}
	\caption{Comparison of changing trend of FID scores versus training epochs between models with and without tessellation for both SW and GW distances. $N$ is set to be 10000 in this experiment. $m=$100, 200, 400 are used for tessellation, and correspondingly the batch sizes are set to 100, 50, 25 for models without tessellation, respectively. }
	\label{fig:fidcompare}
\end{figure}

\begin{figure*}[t!]
	\centering
	\includegraphics[width=\columnwidth]{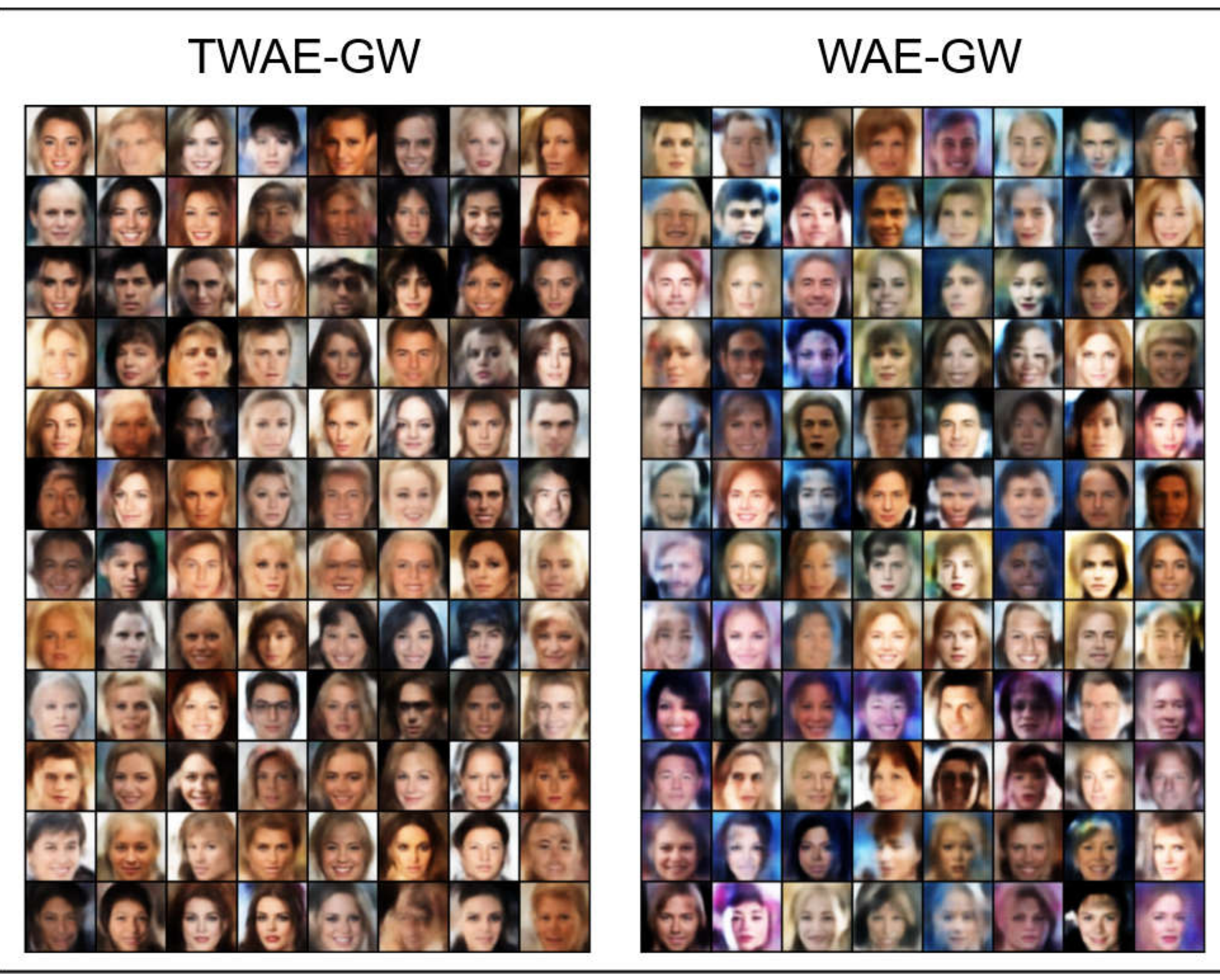}
	\caption{Comparison of generated images using TWAE-GW and WAE-GW with $\lambda$=1 (100 times larger than default). The FID scores of TWAE-GW and WAE-GW are 54.8 and 74.8 respectively.}
	\label{fig:tegw}
\end{figure*}

 Here we show that our tessellation procedure can indeed enhance the performance of non-adversarial methods using existing discrepancy metrics, such as the SW and GW distance. When measure the distance of the two distributions with the GW distance, we treat $P_z$ and $Q_z$ as multivariate Gaussian and ignore the information in the high-order moment. Thus, the approximation is not very good. But with the tessellation technique, actually, we are using a Gaussian mixture distribution with each component in a region to approximate the target distribution. With tessellation, it can be better than the state of the art non-adversarial auto-encoders. Furthermore, for the more accurate discrepancy metrics such as the SW distance, we achieve better performance (Table 4). In Fig. \ref{fig:fidcompare}, we show the downward trends with and without tessellation in the training progress to prove that TWAE has superior generative performance, while keeping the good property of stability. However, for the SW distance, since the decoder of an auto-encoder is only trained with the reconstruction loss, it may not generalize to the ``hole" between the training points. This means increasing the number of regions can not go beyond the generalization ability of the decoder. For instance, the improvements from 200 regions to 400 regions is fewer than that from 100 regions to 200 regions. For the GW distance, when the batch size is smaller than the dimension of the latent space, the computation of $\left(\Sigma_{2}^{1 / 2} \Sigma_{1} \Sigma_{2}^{1 / 2}\right)^{1 / 2}$ in (\ref{eq:Wofgaussian}) is ill-posed. Consequently, the FID score doesn't decrease notably as expected in the case of batch size $=50$ and $25$. Furthermore, TWAE is robust to the hyperparameter $\lambda$. In the case when $\lambda$ is 100 times larger than default (Fig. \ref{fig:tegw}), TWAE-GW can generate distinctly better images (FID=54.8) than without tessellation (FID=74.8).
\begin{table}[t!]
	\centering	
	
	\begin{tabular}{lcc}
		
		\toprule
		distance (batch size) &with tessellation&without tessellation\\
		\midrule
		SW ($100$) &48.5&52.5\\
		SW ($50$)&43.8&51.1\\
		SW ($25$)&43.4&51.5\\
		\midrule
		GW ($100$) &44.5&51.2\\
		GW ($50$)&48.1&50.1\\
		GW ($25$)&57.2&58.6\\
		\bottomrule
	\end{tabular}\label{tab:withoutt}
\caption{Comparison of TWAE with and without tessellation}
\end{table}

\begin{figure*}[t!]
	\centering
	\includegraphics[width=\columnwidth]{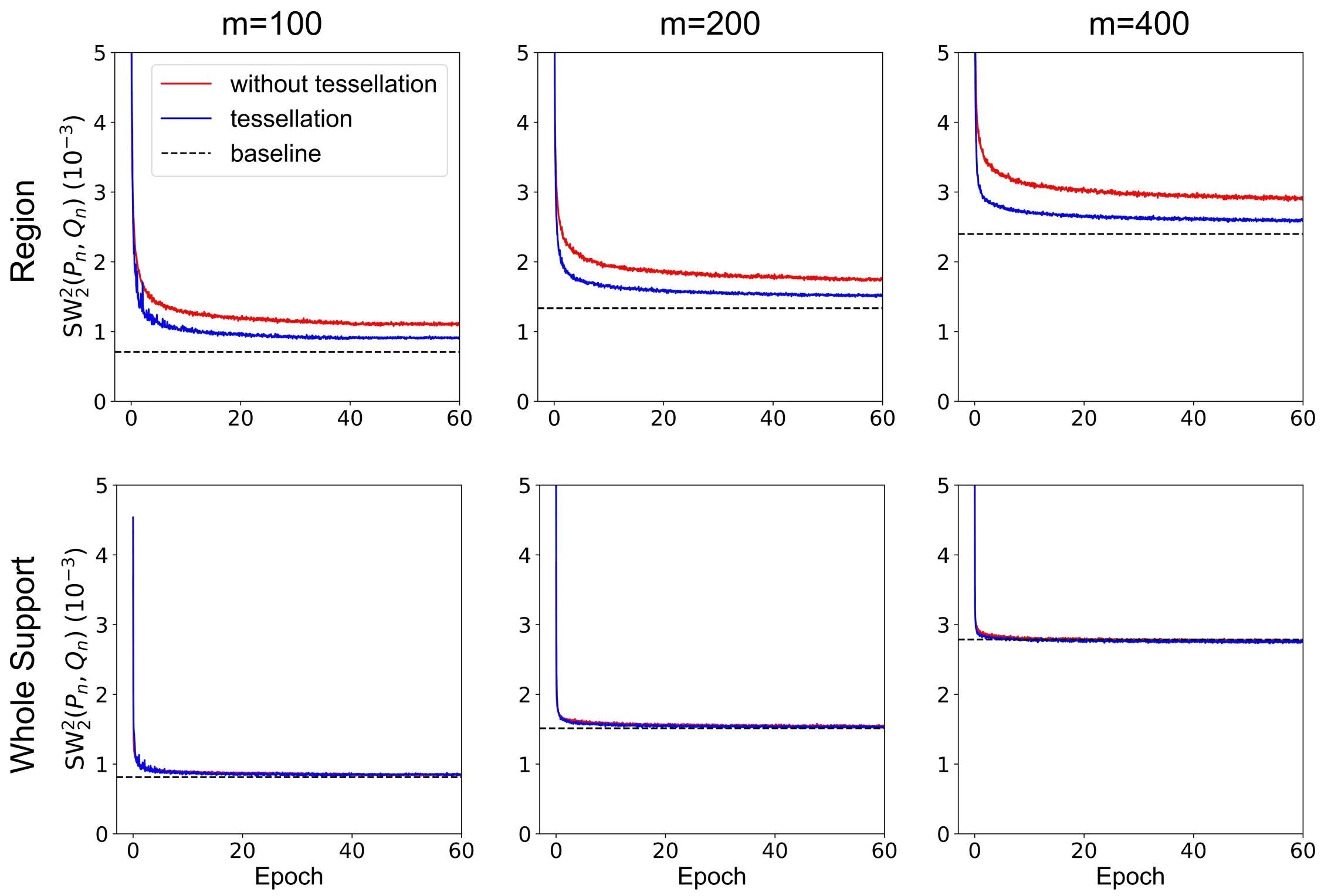}
	\caption{Comparison of the SW distance in the training procedure with tessellation $m=$100, 200, 400 and without tessellation. The baseline is the SW distance of two sets of points sampled from a uniform distribution in the unit ball (the whole support) or in the regions. }
	\label{fig:converge}
\end{figure*}

\par In Fig. \ref{fig:converge}, we show that, at the end of the training procedure, the SW distance can not identify $Q_n$ from $P_n$, i.e., $SW(Q_n,P_n)$ converges to $SW(P_n,P_n^{'})$, where $P_n^{'}$ is sampled from the same $P_z$ as $P_n$. However, in the regions of the whole support, the discrepancy of $P_z$ and $Q_z$ still exists. With tessellation, the SW distance in the regions are closer to sampling from the same distribution, indicating that the tessellation could further reduces the discrepancy.

\subsection{TWAE with other SW distances}

In this subsection, we test TWAE with several recent proposals of SW distances (Max-SW, GSW and DSW) on LSUN-Bedrooms to show its power in enhancing performance of generative auto-encoders. For TWAE, we set $N=10000$ and $m=100$. For GSW, we use circular function to compute the distance. For Max-SW and DSW, the numbers of iterations to find the optimal projection and the optimal distribution of projection are both set to 10. The numbers of projections are all set to 1000 for SW, DSW and GSW respectively.

\begin{table}[hpbt]
	\centering
	
	\setlength{\tabcolsep}{7mm}{
		\begin{tabular}{lclc}
			\toprule
			\textbf{Model} & \textbf{FID} & \textbf{Model} & \textbf{FID}\\
			\midrule
			TWAE-SW & 193.2 & WAE-SW & 205.4\\
			TWAE-GSW & 193.7 & WAE-GSW & 214.1\\
			TWAE-DSW & 196.3 & WAE-DSW & 210.3 \\
			\bottomrule
	\end{tabular}}\label{table:comparison_SW}
	\caption{Performance comparison of the SW, GSW and DSW distances on the LSUN-Bedrooms dataset.}
\end{table}

As we expected, the tessellation technique can enhance the performance of auto-encoders with SW, GSW and DSW uniformly (Table 5). The performance of TWAE-SW, TWAE-GSW and TWAE-DSW are very similar, while the performance of WAE-GSW and WAE-DSW is even slightly worse than that of WAE-SW. This is because DSW is designed for GAN, in which the latent distributions are complex and anisotropic. A few of projection samples in SW are more important than the rest. DSW finds an optimal distribution of important projection samples, which leads to better performance in GAN. Also, GSW is designed to model the irregular support shape of the latent distribution. However, in the setting of auto-encoder, when the prior latent distribution is uniform in a unit ball, each projection samples contributes equally in SW and the support shape of latent distribution is regular. Thus, SW is better than GSW and DSW for WAE. It should be noted that Max-SW finds the most important projection sample, but ignores the rest which is still important due to the isotropic latent distribution. Thus, WAE and TWAE with Max-SW fail to learn the distribution and don’t converge in this test.

\section{Discussion and Conclusion}
\par In this paper, we propose a novel non-adversarial generative framework TWAE, which designs batches according to data similarity instead of random shuffling, and optimizes the discrepancy locally. It shows very competitive performance to an adversarial generative model WAE-GAN, while sharing the stability of other non-adversarial ones. It is very flexible and applicable to different discrepancy metrics to enhance their performance. To our knowledge, TWAE is the first generative model to design batches and optimize with non-identical distributions. To this end, we use a computational geometry technique CVT, which is often used in three-dimensional modeling, and develop a new optimization method to deal with such non-identical batches. TWAE can generate images of higher quality in terms of FID score with relatively more regions when the computing resource is adequate.

\par TWAE is designed to learn the data distribution in the latent space learned by an auto-encoder model, instead of the original space $(d>1000)$ of data (e.g., images). Generally, the distribution of data concentrates near a low-dimension manifold, so the similarity should be measured by the Riemann metric on the manifold rather than the Euclidean metric. However, construction of the Riemann metric in high dimensional space without neural network is hard. Thus, we suggest to tessellate the latent space to approximate the target distribution better. Here we suggest to use the uniform distribution but not the i.i.d. Gaussian as the prior distribution of the latent space. The reasons for this are threefold: 1) for a uniform distribution, the probability of a region $P(V_i)$ is corresponding to its volume. It is convenient to conduct tessellation with equal-weighted sampling; 2) uniform distribution is isotropic when restricted to a region. While computing the SW distance, projections of different directions have useful information because the distribution is isotropic; 3) when the points obey uniform distribution, we can use the Euclidean metric to measure the similarity of two points.

\par Since the decoder $\psi$ is trained on $Q_z$ rather than $P_z$, the quality of generated images may not be as good as that of GAN. In some situations, people care about generating more than encoding. It is nontrival to generalize the tessellation technique to GAN. The reason for this is two-fold: 1) The adversarial mechanism is unstable and sensitive to noise, thus the variance induced by such designed batches may impede the optimization process of GAN; 2) In GAN, there is no encoder to extract high-level representation of data, which makes it difficult to cluster the data into batches according to their similarity. Nevertheless, it will be valuable to develop a technique analogous to tessellation that can enhance the performance of GAN.

\par In TWAE, since the supports of distributions of different batches are disjoint, the model does not forget the information in passed batches when learn with a new batch. However, neural network tends to forget the knowledge of previously learned tasks as information relevant to the current task is incorporated. This phenomenon is termed catastrophic forgetting. For instance, in the situations of online machine learning, data becomes available in sequential order. So the distribution of each batch may change, and previously learned knowledge might lose. Numerical experiments showed that our optimization method can deal with non-identical batches, i.e., learning from the current batch without forgetting the former batches. Can techniques in catastrophic forgetting help to further reduce the gap of the SW distance in the regions (Fig. \ref{fig:converge})? Or can our non-identical batch optimization help to overcome the catastrophic forgetting? They will be valuable questions worthing further studying.

\par As mentioned above, the numbers of minimal vectors of $E_8$-lattice and Leech lattice for 8- and 24-dimension cases are 240 and 196560 respectively. So the data we have actually can not fill the latent space when the dimension is very high. Some bad images will be generated when we randomly sample in the latent space due to the lack of data points. Unfortunately, there is no criterion to judge whether the sampled point in the latent space can generate a good image. In the future, how to build the statistics to evaluate the quality of the generated images and find the well-learned region in the latent space is an important topic.

\acks{
This work has been partially supported by the National Key R\&{D} Program of China [2019YFA0709501]; the National Natural Science Foundation of China [61621003]; National Ten Thousand Talent Program for Young Top-notch Talents; CAS Frontier Science Research Key Project for Top Young Scientist [QYZDB-SSW-SYS008]. }


\vskip 0.2in
\bibliography{reference}
\end{document}